\documentclass[11pt,a4paper]{article}

\usepackage{booktabs} %
\usepackage{amsmath}
\usepackage{amssymb} %
\usepackage{subcaption}  %
\usepackage{acronym} %
\usepackage{threeparttable}  %
\usepackage[colorinlistoftodos]{todonotes}  %

\usepackage[nocustomfonts]{nxai}
\newcommand{\affmark}[1]{\raisebox{0.55ex}{\fontsize{7}{7}\selectfont #1}}

\nxaisetalogo{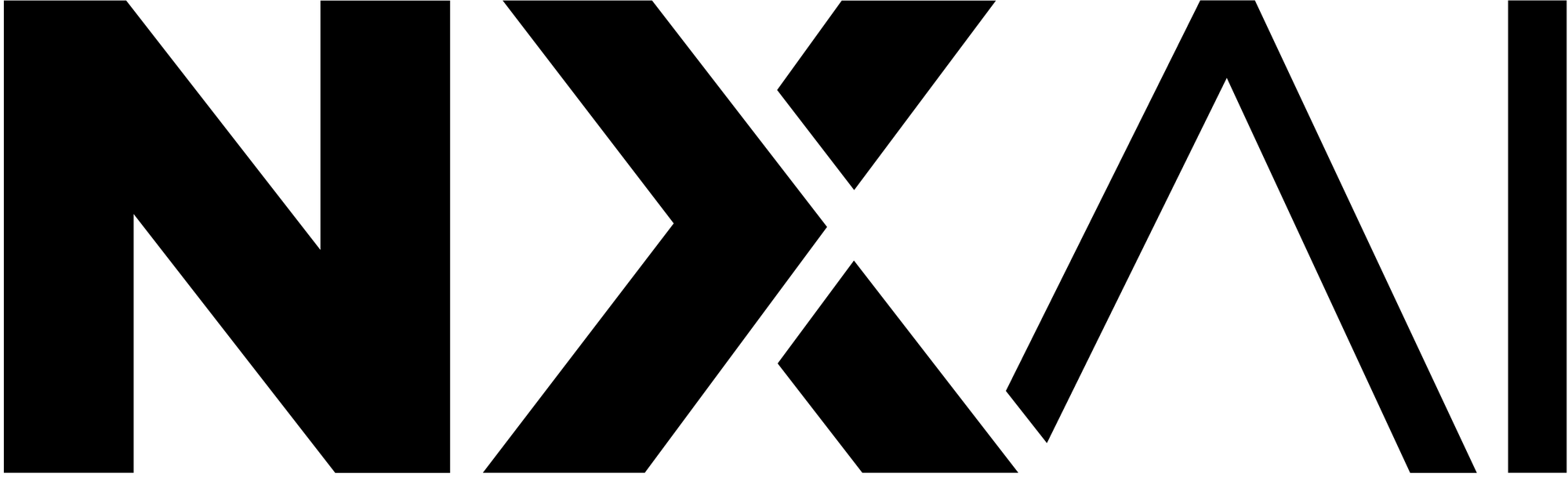} %
\nxaisetlogoheight{4mm} %
\nxaisetdate{July 6, 2026}

\title{Effective Distillation to Hybrid xLSTM Architectures}
\author{
    Lukas Hauzenberger\affmark{1,2,*}\and
    Niklas Schmidinger\affmark{1,2,*}\and
    Thomas Schmied\affmark{1,2}\and
    Anamaria-Roberta Hartl\affmark{2}\and
    David Stap\affmark{1}\and
    Pieter-Jan Hoedt\affmark{2}\and
    Maximilian Beck\affmark{1,2}\and
    Sebastian B{\"o}ck\affmark{1}\and
    G{\"u}nter Klambauer\affmark{1,2}\and
    Sepp Hochreiter\affmark{1,2}
}
\nxaisetaffiliations{%
\affmark{1}NXAI
\affmark{2}Johannes Kepler University Linz
\affmark{*}Equal contribution.
}

\newacro{llm}[LLM]{large language model}
\newacro{swa}[SWA]{sliding window attention}
\newacro{rnn}[RNN]{recurrent neural network}
\newacro{mlp}[MLP]{multi-layer perceptron}
\newacro{rope}[RoPE]{rotary position embedding}
\newacro{mse}[MSE]{mean-squared error}
\newacro{ce}[CE]{cross-entropy}
\newacro{kl}[KL]{Kullback-Leibler divergence}
\newacro{ppl}[PPL]{perplexity}
\newacro{peft}[PEFT]{parameter-efficient fine-tuning}
\newacro{fft}[FFT]{full fine-tuning}
\newacro{lora}[LoRA]{low-rank adaptation}
\newacro{ttft}[TTFT]{time to first token}
\newacro{oom}[OOM]{out of memory}
\newacro{rl}[RL]{reinforcement learning}
\newacro{ssm}[SSM]{state space model}
\newacro{fsdp}[FSDP]{fully sharded data parallel}
\newacro{kv}[KV]{key-value}

\usepackage{pifont}
\newcommand\cmark{\ding{51}}
\newcommand\xmark{\ding{55}}

\newcommand\llamaStudent{\textsc{xLSTM-Llama3.1-8B}}
\newcommand\llamaStudentIT{\textsc{xLSTM-Llama3.1-8B-IT}}
\newcommand\llamaStudentITGeneralist{\textsc{xLSTM-Llama3.1-8B-IT-Generalist}}

\newcommand\qwenStudentIT{\textsc{xLSTM-Qwen2.5-7B-IT}}
\newcommand\qwenStudentITGeneralist{\textsc{xLSTM-Qwen2.5-7B-IT-Generalist}}

\newcommand\olmoStudent{\textsc{xLSTM-Olmo3-7B}}

\newcommand\llamaTeacher{\textsc{Llama3.1-8B}}
\newcommand\llamaTeacherIT{\textsc{Llama3.1-8B-IT}}
\newcommand\llamaThreeTeacherIT{\textsc{Llama3-8B-IT}}

\newcommand\qwenTeacher{\textsc{Qwen2.5-7B}}
\newcommand\qwenTeacherIT{\textsc{Qwen2.5-7B-IT}}

\newcommand\olmoTeacher{\textsc{Olmo3-7B}}

\newcommand\baselineLolcats{\textsc{LoLCATs}}
\newcommand\baselineMambaLlama{\textsc{Mamba-in-Llama}}
\newcommand\baselineQrwkv{\textsc{QRWKV7-7B-IT}}
\newcommand\baselineQrwkvBase{\textsc{QRWKV6-7B}}

\nxaiabstract{%
There have been numerous attempts to distill quadratic attention-based \acp{llm} into sub-quadratic linearized architectures. 
However, despite extensive research, such distilled models often fail to match the performance of their teacher \acp{llm} on various downstream tasks.
We set out the goal of \emph{lossless distillation}, which we define in terms of 
tolerance-corrected \emph{Win-and-Tie rates} between student and teacher on sets of tasks.
To this end, we introduce an effective distillation pipeline for xLSTM-based students.
We propose an additional merging stage, where individually linearized experts are combined into a single model.
We show the effectiveness of this pipeline by distilling base and instruction-tuned models from the Llama, Qwen, and Olmo families.
In many settings, our xLSTM-based students recover most of the teacher's performance, and even exceed it on some downstream tasks.
Our contributions are an important step towards more energy-efficient and cost-effective replacements for transformer-based \acp{llm}.
}

\begin{document}
\maketitle

\section{Introduction}
\label{sec:intro}

Current \acp{llm} require enormous computational resources because their attention mechanisms \citep{vaswani2017attention, touvron2023llama, bai2023qwen,openai2025gpt-oss} scale quadratically with context length. 
As a result, these models are energy-intensive and costly to deploy. 
To address these limitations, many works distill \citep{schmidhuber1991neural,hinton2015distilling} \acp{llm} into linearized, attention-free, or more generally 
sub-quadratic architectures \citep{wang2024mamba, bick2024transformers,wang2025m1,zhang2025lolcats, lan2025liger, goldstein2025radlads}.
The efficient inference of sub-quadratic distilled \acp{llm} 
makes them favorable drop-in replacements, 
if they match their teachers across diverse tasks.

\begin{figure}[htp]
    \centering
    \begin{subfigure}{.48\linewidth}
        \includegraphics[width=\linewidth]{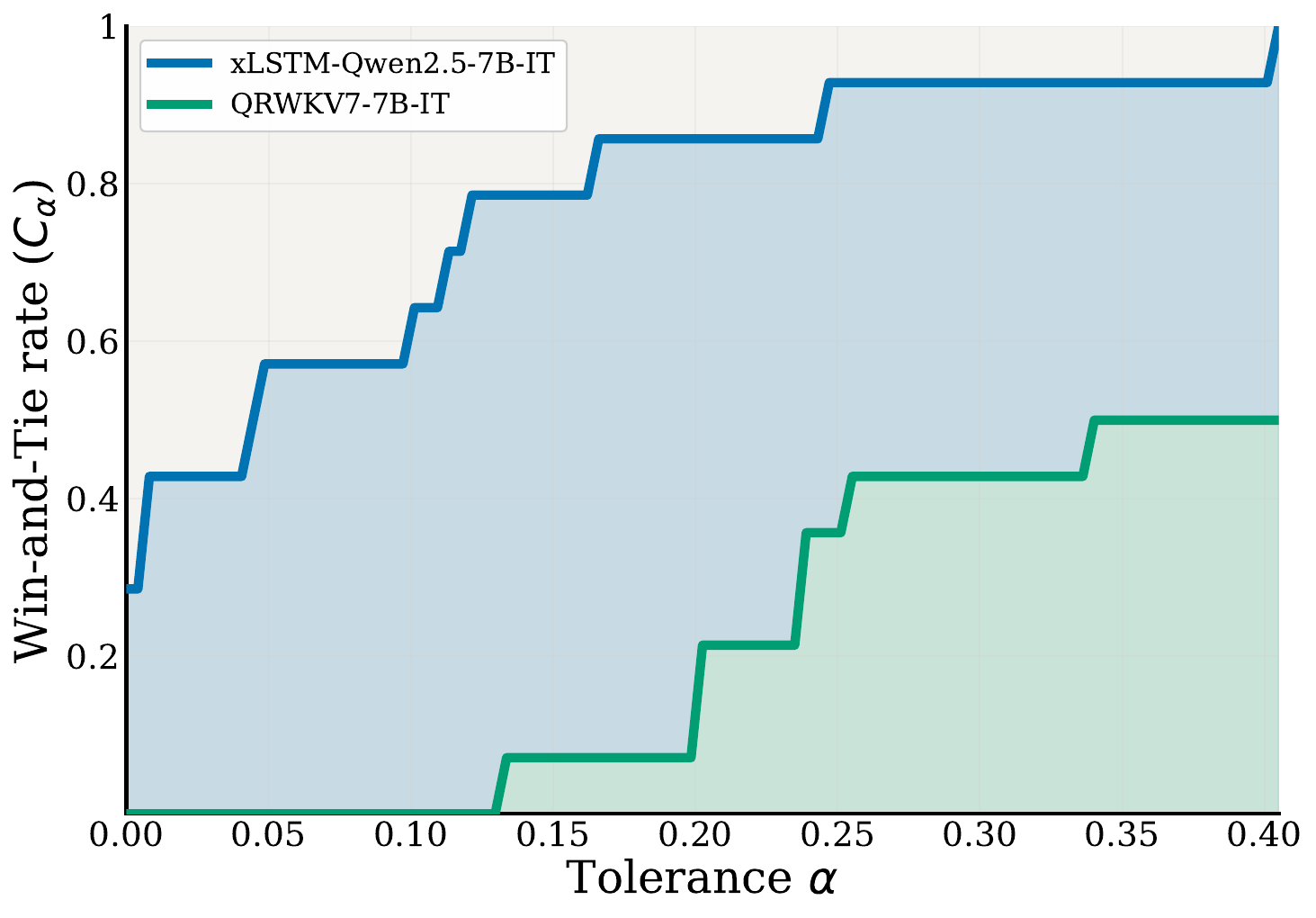}
    \end{subfigure}%
    \hfill
    \begin{subfigure}{.48\linewidth}
        \includegraphics[width=\linewidth]{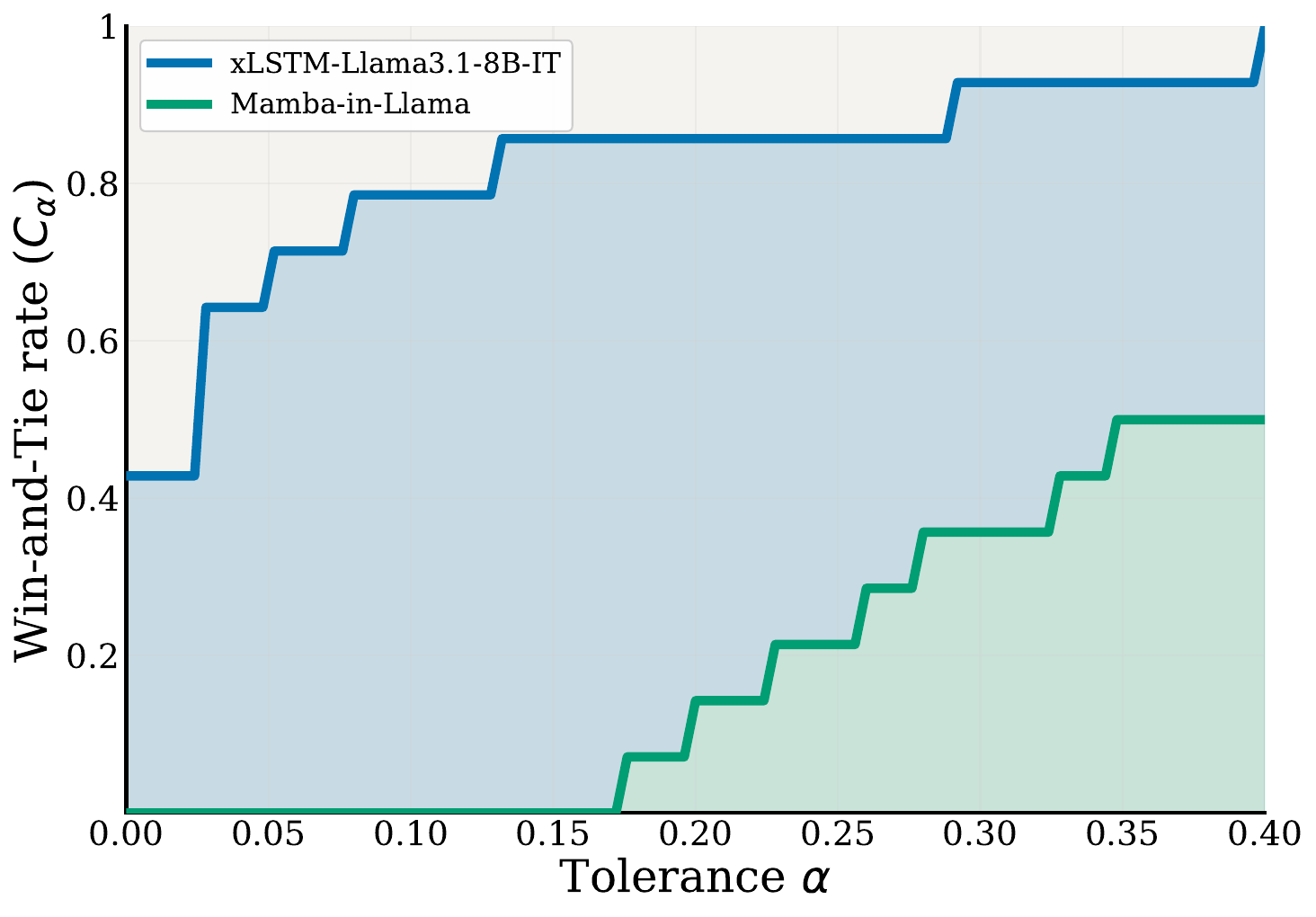}
    \end{subfigure}
    \caption{
        Win-and-Tie rate ($C_\alpha$) curves of our distilled \qwenStudentIT  \textbf{(left)} and \llamaStudentIT  \textbf{(right)} in comparison against  the best sub-quadratic baseline across generation benchmarks spanning math, code, STEM, and chat domains.
        Higher is better.
    }
    \label{fig:pareto_qwen_main}
\end{figure}

\newpage

Recent post-training linearization has coalesced around a handful of 
sub-quadratic sequence mixer designs to \emph{substitute} full softmax attention layers and a small set of recurring distillation techniques.
{LoLCATs} \citep{zhang2025lolcats} and {Liger} \citep{lan2025liger} implement intra-layer hybrids that couple linear attention variants with \ac{swa}, RADLADS \citep{goldstein2025radlads} adapt {RWKV-6} \citep{peng2024eagle} and {RWKV-7} \citep{peng2025rwkv7} for the distillation setting, and {Llamba} \citep{bick2025llamba} converts layers to {Mamba-2} state-space mixers \citep{dao2024transformers}. 
For linearization, supervision typically involves hidden-state and logit alignment on small subsets of general web-text mixtures or instruction datasets.
In contrast, token budgets for conventional \ac{llm} pre-training range from tens of billions to trillions of tokens, rendering linearization orders of magnitude more token-efficient than training from scratch. 
Therefore, linearization is an attractive fine-tuning regime for both exploring novel linear attention designs and lowering the deployment cost of Transformer-based models. 
However, existing linearization attempts have not yet achieved effective distillation. 
While linearized models often match the teacher on language understanding or knowledge benchmarks, they fall short on harder generative evaluations that probe the student's mathematical reasoning or code synthesis abilities (see Figures~\ref{fig:downstream-evals-base}b and~\ref{fig:downstream-evals-instruct}).
These outcomes highlight limitations of existing distillation procedures, architectures, and evaluation protocols (see Appendix~\ref{sec:related-work-linearization} for an overview of prior work).

\textbf{xLSTM as a powerful linear alternative for \acp{llm}.}
Recently, modern recurrent architectures, such as xLSTM \citep{beck2024xlstm}, Gated Delta Networks \citep{yang2024gdn}, and Mamba \citep{gu2024mamba}, have emerged as competitive linear-complexity alternatives to Transformers in language \citep{beck2025xlstm7b}, computer vision \citep{alkin2025vision,poppel2025plstm}, biological modeling \citep{schmidinger2025bio-xlstm}, decision-making \citep{schmied2025large}, and time series \citep{auer2025tirex}.
Concurrently, specialized kernels enable efficient chunkwise-parallel training for 
linear \acp{rnn} and xLSTM, substantially improving throughput on high-end accelerators \citep{beck2025tiled}.
Recent scaling-law analyses further indicate that xLSTM maintains competitive advantages as training and inference contexts grow, positioning it as a strong foundation for efficient long-context models \citep{beck2025scaling}. 
We hybridize xLSTM with sparse attention by combining an mLSTM with a synchronous \ac{swa} path and sink tokens using learned gates.
Conceptually, this is related to recent attention hybrids that blend quadratic \ac{kv} memory with linear fast-weight memory \citep{irie2025blending}.

\textbf{Contributions.}
To rigorously assess whether linearized students can serve as drop-in replacements, we formalize a reliability criterion via the \emph{Win-and-Tie rate} $C_\alpha$, which measures how broadly the student recovers teacher-level performance across benchmarks.
Using this criterion, we show that prior linearization approaches often preserve language understanding but fall short on harder, free-form generation tasks.
To close this gap, we introduce a linearization pipeline that replaces quadratic softmax attention with an efficient mLSTM--\ac{swa} hybrid.
In our linearization pipeline we introduce a merging stage, where domain-specialized students are distilled independently and consolidated afterwards.
In this sense, we demonstrate that linearization can be made modular: linearized models can be consolidated through simple weight-space merging \citep{wortsman2022model}.
The resulting merge of distilled xLSTM students closes the performance gap on free-form generation tasks and consistently dominates existing linearization methods across tolerance levels on $C_\alpha$.

\section{Background}
\label{sec:background}

\begin{figure*}
  \centering
  \includegraphics[width=\linewidth]{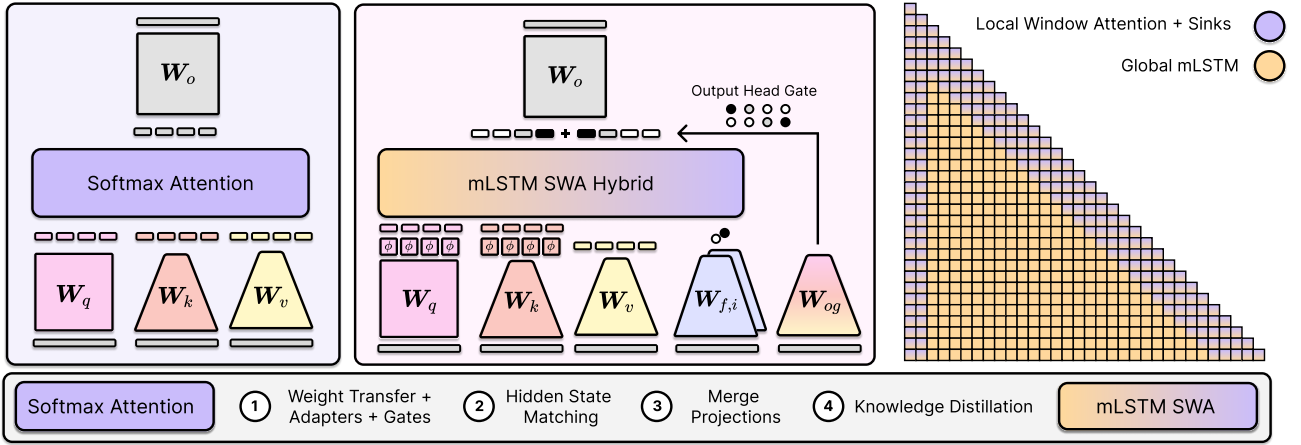}
  \caption{
    Illustration of our hybrid method consisting of mLSTM, sliding-window attention, and sink tokens.
    Our approach comprises 4 primary steps: \textbf{(1)} transfer the original teacher weights to the student and introduce adapters and gates, \textbf{(2)} hidden-state matching, \textbf{(3)} subsequent merging of query and key projections, and \textbf{(4)} knowledge distillation.
  }
  \label{fig:overview}
\end{figure*}

\textbf{Softmax attention and Transformers.}
The impressive capabilities of Transformer-based \acp{llm} are largely attributed to the effectiveness of the underlying softmax-attention mechanism \citep{vaswani2017attention}, which enables fine-grained modeling of long-range dependencies. 
At each time step $t$, an attention layer receives an input $\boldsymbol{x}_t \in \mathbb{R}^{d}$ and projects it to a query $\boldsymbol{q}_t$, key $\boldsymbol{k}_t$, and value $\boldsymbol{v}_t$ via learned linear maps $\boldsymbol{W}_q,\boldsymbol{W}_k \in \mathbb{R}^{d \times d_{qk}}$ and $\boldsymbol{W}_v \in \mathbb{R}^{d \times d_v}$:
\begin{align}
    \boldsymbol{q}_t &= \boldsymbol{x}_t\boldsymbol{W}_q&
    \boldsymbol{k}_t &= \boldsymbol{x}_t\boldsymbol{W}_k&
    \boldsymbol{v}_t &= \boldsymbol{x}_t\boldsymbol{W}_v
\end{align}
so that $\boldsymbol{q}_t,\boldsymbol{k}_t \in \mathbb{R}^{d_{qk}}$ and $\boldsymbol{v}_t \in \mathbb{R}^{d_v}$.
To avoid recomputation, \ac{kv} caches are maintained whose sizes grow with time, $\boldsymbol{K}_t \in \mathbb{R}^{t \times d_{qk}}$ and $\boldsymbol{V}_t \in \mathbb{R}^{t \times d_v}$, updated by concatenation (denoted as $[\,]$) along the time dimension:
\begin{align}
    \boldsymbol{K}_t &= [\,\boldsymbol{K}_{t-1}\;\; \boldsymbol{k}_t\,] &
    \boldsymbol{V}_t &= [\,\boldsymbol{V}_{t-1}\;\; \boldsymbol{v}_t\,]
    \label{eq:kv-cache}
\end{align}
The output is then read from memory using \emph{scaled softmax attention}:
\begin{equation}
    \boldsymbol{h}_t = \,\operatorname{softmax}\;\Bigl(\tfrac{1}{\sqrt{d_{qk}}}\, \boldsymbol{q}_t\boldsymbol{K}^{\top}_t\Bigr)\boldsymbol{V}_t
    \label{eq:scaled-softmax-attn}
\end{equation}
For a given query $\boldsymbol{q}_t$ at $t$, the dot product between the query and all stored keys $\boldsymbol{K}_t$ up to $t$ is computed.
Subsequently, $\operatorname{softmax}$ is applied over time steps, and the resulting per-position attention scores are used to compute a weighted average of the stored values $\boldsymbol{V}_t$.
During training and context encoding, each query in a sequence of length $T$ is compared with every key, incurring $O(T^2)$ time.
During autoregressive inference, \ac{kv} pairs are appended to the cache.
At step $t$, the attention readout time and the cache size are $O(t)$. 
Although linear in $t$, the cache footprint scales with network depth and heads, and the cache must be read at every step, implying $O(t)$ memory-bandwidth cost. 
For long contexts, this becomes a dominant system bottleneck on modern accelerators, constraining batch size and throughput and increasing latency.

\textbf{Sparse and Sliding-window attention.} 
To mitigate the training and inference costs of full softmax attention, many \acp{llm} adopt sparse attention patterns in which each head attends only to a subset of past positions \citep{child2019generating, beltagy2020longformer, zaheer2020big, yuan2025native}. 
A widely used special case is \acf{swa}, which restricts each query to attend to a fixed-length band of its immediate token history.
\Ac{swa} evicts keys and values outside the last $W$ steps:
\begin{align}
    \boldsymbol{K}_t^W &= \bigl[\,\boldsymbol{K}_{t-1}^{W-1}\;\; \boldsymbol{k}_t\,\bigr] &
    \boldsymbol{V}_t^W &= \bigl[\,\boldsymbol{V}_{t-1}^{W-1}\;\; \boldsymbol{v}_t\,\bigr]
    \label{eq:kv-cache-swa}
\end{align}
where $\boldsymbol{K}_t^W = \boldsymbol{K}_{\max(1, t - W + 1):t} \in \mathbb{R}^{\min(t, W) \times d_{qk}}$ and $\boldsymbol{V}_t^W = \boldsymbol{V}_{\max(1, t - W + 1):t} \in \mathbb{R}^{\min(t, W) \times d_{v}}$.
The maximum cache length of \ac{swa} therefore never exceeds $W$, while the core attention computation remains unchanged (see Equation~\ref{eq:scaled-softmax-attn}). 
For sequences of length $T$, training and prefill of \ac{swa} can be implemented in linear $O(TW)$ time instead of $O(T^2)$ for full softmax attention.
Consequently, during autoregressive decoding, both the computational and memory complexities of \ac{swa} are independent of the global sequence length.
In Appendix~\ref{appendix:receptive-field-of-swa} we discuss the effective receptive field of \ac{swa}.

\newpage

\textbf{Linear attention} replaces the exponential kernel of softmax attention $\kappa_{\exp}(\boldsymbol{q},\boldsymbol{k})=\exp\!\big(\boldsymbol{q}^\top \boldsymbol{k}/\sqrt{d_{qk}}\big)$ with a finite-dimensional feature map $\phi:\mathbb{R}^{d_{qk}}\!\to\!\mathbb{R}^{d_{qk}}$ such that $\kappa_{\phi}(\boldsymbol{q},\boldsymbol{k})=\phi(\boldsymbol{q})^\top \phi(\boldsymbol{k})$ \citep{katharopoulos2020transformers}.
This factorization enables two efficient implementations of causal attention: a chunkwise-parallel form for training and context encoding and a strictly recurrent form for stepwise decoding \citep[see e.g.][]{yang2024gated}.
Switching between these views enables prefill and training in linear time and constant-memory generation. 
In the recurrent view, we maintain a per-head \ac{kv} state $\boldsymbol{S}_t\in\mathbb{R}^{d_{qk}\times d_v}$ that accumulates prefix statistics via rank-1 outer-product updates, together with an optional normalizer $\boldsymbol{z}_t\in\mathbb{R}^{d_{qk}}$:
\begin{align}
    \boldsymbol{S}_{t} &= \boldsymbol{S}_{t-1}+\phi(\boldsymbol{k}_{t}) \otimes \boldsymbol{v}_{t} \\
    \boldsymbol{z}_{t} &= \boldsymbol{z}_{t-1}+\phi(\boldsymbol{k}_{t})
    \label{eq:linattn-recurrent}
\end{align}
$\otimes$ denotes the outer product.
Given a query $\boldsymbol{q}_t$, we perform a normalized read from the current state:
\begin{equation}
    \boldsymbol{h}_{t} = \frac{\phi(\boldsymbol{q}_{t}) \boldsymbol{S}_{t}}{\phi(\boldsymbol{q}_{t})\boldsymbol{z}_{t}}
    \label{eq:linattn-read}
\end{equation}
Here, $\boldsymbol{q}_t,\boldsymbol{k}_t\in\mathbb{R}^{d_{qk}}$ and $\boldsymbol{v}_t\in\mathbb{R}^{d_v}$. 

\textbf{mLSTM}. 
Inspired by the LSTM cell \citep{hochreiter1997lstm}, the mLSTM \citep{beck2024xlstm} augments the linear attention update with three data dependent gates that control distinct aspects of the update: $\boldsymbol{w}_i \in \mathbb{R}^{d \times 1}$, $\boldsymbol{w}_f \in \mathbb{R}^{d \times 1}$, $\boldsymbol{W}_{og} \in \mathbb{R}^{d \times d_{v}}$ where the input gate activations $i_t = \operatorname{exp}(\boldsymbol{x}_t \boldsymbol{w}_i)$ set the strength of the new \ac{kv} write, the forget gate activations $f_t = \sigma(\boldsymbol{x}_t \boldsymbol{w}_f)$ decay the accumulated state, and the output gate activations $\boldsymbol{o}_t = \sigma(\boldsymbol{x}_t \boldsymbol{W}_{og})$ modulate the readout:
\begin{align}
    \boldsymbol{S}_t &= f_t\,\boldsymbol{S}_{t-1}+i_t\,\phi(\boldsymbol{k}_t)\otimes \boldsymbol{v}_t \\\boldsymbol{z}_t &= f_t\,\boldsymbol{z}_{t-1}+i_t\,\phi(\boldsymbol{k}_t)
    \label{eq:mlstm-recurrent}
\end{align} 
Numerical stabilization for the exponential input gate is omitted for simplicity.
A query then performs a normalized read, and the output gate modulates the retrieved value:
\begin{equation}
    \boldsymbol{\hat h}_t = \boldsymbol{o}_t \odot \frac{\phi(\boldsymbol{q}_{t}) \boldsymbol{S}_{t}}{\phi(\boldsymbol{q}_{t}) \boldsymbol{z}_{t}}
    \label{eq:mlstm-read}
\end{equation}

\section{xLSTM distillation pipeline}
\label{sec:method}

In this work, we propose a distillation pipeline for creating efficient \acp{llm}, 
substituting full softmax attention with a sub-quadratic attention proxy. 
The core of our method involves replacing the standard self-attention mechanism 
in a pre-trained \acp{llm} with a hybrid attention block that combines \ac{swa} with mLSTM\footnote{We use the xLSTM\textsubscript{[1:0]} 
configuration, which employs xLSTM blocks with mLSTM cells only.} \citep{beck2024xlstm} via data-dependent gating.

\subsection{Architecture \& student initialization}

We use a pretrained causal Transformer-based \ac{llm} as the \emph{teacher} model, 
similar to prior work \citep{zhang2025lolcats}.
The \emph{student} adopts the same high-level architecture design as the teacher, 
while replacing every multi-head attention block with a hybrid of \ac{swa} and mLSTM.
This allows us to recycle the parameters of the original embedding and attention layers and the \ac{mlp} blocks.
The fundamental motivation for our hybrid approach is to combine the strengths 
of two distinct and efficient sequence-mixing paradigms: the local context capturing ability of \ac{swa} and the linear complexity of mLSTM.
Both components operate in parallel, and their outputs are dynamically fused using a learned, data-dependent gate.

\newpage

\textbf{mLSTM adaptations}.
Recent instantiations of gated linear operators replace 
the classical normalizer state with normalization layers such as LayerNorm \citep{sun2023retentive, yang2024gated, beck2025tiled}.
In the linearization setting, we observe that adding normalization immediately before the output projections degrades student-teacher alignment.
Similar observations have been made in \cite{bick2025llamba}.
For this reason, we opt for the original normalizer design (cf. Equation~\ref{eq:mlstm-read}) without normalization layers.

Instead of one output gate per channel, as in the original mLSTM, we use per-head 
scalar output gates to keep the parameter count closer to that of the teacher model.
Furthermore, we found that using a concatenation of the head inputs over the feature dimension
$\bigl[\,\boldsymbol{q}_{t}\;\;\boldsymbol{k}_{t}\;\;\boldsymbol{v}_t\,\bigr]$ instead of the input activations $\boldsymbol{x}_t$ at time $t$ provides a better input signal for the output gate projections $\boldsymbol{W}_{og}$.
Due to the strictly linear nature of the combination of $\boldsymbol{W}_q$, $\boldsymbol{W}_k$, $\boldsymbol{W}_v$ projections and $\boldsymbol{W}_{og}$, this can be merged to a single linear projection with input $\boldsymbol{x}_t$ at any stage.
We augment the query and key inputs to the mLSTM with head-wise feature maps, applying softmax over the feature dimension as the activation function \citep{zhang2024hedgehog}.

\textbf{Attention hybridization \& data-dependent output mixing.}
We combine mLSTM and sparse attention into a single unified attention block, 
similar to \citet{zhang2025lolcats} and \citet{dong2025hymba}, rather than alternating both operators at every layer. %
We opt for a sparse attention pattern using \ac{swa} over the most recent token history and four initial tokens per sequence to preserve attention sinks, similar to \citet{xiao2024efficient}.
The combination of \ac{swa} and sink tokens enables both efficient \ac{kv} cache compression and a good initial approximation of full softmax attention.
For a discussion on attention sinks, we refer to Appendix~\ref{appendix:attn-sinks-perspective}.
In Section~\ref{sec:ablations}, we demonstrate that all three components are critical for strong performance. Moreover, in Appendix~\ref{sec:related-work-hybrid-archs}, we contextualize our architectural design relative to contemporary hybrid linear-attention architectures.

For a given input batch, we compute query, key, and value activations and apply \acp{rope} \citep{su2024roformer}. 
The output of the \emph{local} \ac{swa} + sink branch is computed using a sparse attention kernel \citep{dong2024flex}.
For the \emph{global} mLSTM branch, we transform queries and keys with our head-wise feature maps $\phi$ and pass them together with input and forget gate activations to the mLSTM cell.
Finally, the output gate produces a sigmoid-bounded scalar per head that modulates the global mLSTM against the local \ac{swa} + sink outputs, similar to \citet{yuan2025native} and \citet{irie2025blending}:
\begin{equation}
    \begin{aligned}
    \boldsymbol{\hat h}_t = o_t \operatorname{mLSTM}\bigl(\boldsymbol{q}_t\bigr) + (1 - o_t) \operatorname{SWA}\bigl(\boldsymbol{q}_t\bigr)
    = o_t \frac{\phi(\boldsymbol{q}_t) \boldsymbol{S}_t}{\phi(\boldsymbol{q}_t) \boldsymbol{z}_t} + \left(1-o_t\right) \operatorname{sm}\left(\frac{\boldsymbol{q}_t{\boldsymbol{K}_t^W}^\top}{\sqrt{d_{qk}}}\right)\boldsymbol{V}_t^W,
    \end{aligned}
    \label{eq:mlstm-hybrid}
\end{equation}
where $\operatorname{sm}$ is used as a short form for softmax.
This simple yet effective combination of mLSTM and \ac{swa} yields a harmonic interplay between modeling short and long-term dependencies.

\subsection{Linearization fine-tuning}

\textbf{Linearization stage~I: layer-wise hidden-state alignment.}
Following prior linearization work (see Appendix~\ref{sec:related-work-linearization}), 
we first align the per-layer representations of the student to the attention outputs of the teacher using a \ac{mse} objective.
For each layer $\ell$ and time step $t$, let $\boldsymbol{h}^{(\ell)}_t=\operatorname{SoftmaxAttention}(\boldsymbol{q}^{(\ell)}_t,\boldsymbol{K}^{(\ell)}_t,\boldsymbol{V}^{(\ell)}_t)$ denote the teacher’s attention output and let $\boldsymbol{\hat h}^{(\ell)}_t$ denote the corresponding student hidden state as defined in Equation~\eqref{eq:mlstm-hybrid}.
The layer-wise objective is:
\begin{equation}
    \min_{\boldsymbol{\theta}_\ell}\;
    \big\|\,\boldsymbol{h}^{(\ell)}_t-\boldsymbol{\hat h}^{(\ell)}_t\,\big\|_2^2,
    \label{eq:align}
\end{equation}
where $\boldsymbol{\theta}_\ell$ denotes the newly introduced parameters, i.e., the parameters of the head-wise feature maps and gate projections.
The embedding and \ac{mlp} weights from the teacher are frozen in this stage.
The full batch loss is then computed as the sum of Equation~\eqref{eq:align} over layers and time.

\newpage

\textbf{Linearization stage~II: sparse knowledge distillation.}
Following the hidden-state alignment stage, we unfreeze all student parameters $\boldsymbol{\theta}$ and fine-tune end-to-end.
This stage can be viewed as cross-architecture knowledge distillation from a Transformer teacher into a recurrent/hybrid student, conceptually related to prior work on transferring inductive biases between heterogeneous model families such as Transformers and LSTMs \citep{abnar2020transferring}.
The objective for this stage interpolates between next-token prediction and matching the teacher distribution via the \ac{kl}:
\begin{equation}
    \begin{aligned}
        \min_{\boldsymbol{\theta}} \Bigl\{
            -\sum_{t=1}^{T} \gamma\,\log p_{\boldsymbol{\theta}}\left(y_{t} \mid \mathbf{x}_{1:t}\right) + \beta \,\operatorname{KL}\left[p_T^{(k)}\left(\cdot \mid \mathbf{x}_{1:t}\right) \mathbin{\big\|} p_{\boldsymbol{\theta}}^{(k)}\left(\cdot \mid \mathbf{x}_{1:t}\right)\right]
        \Bigr\},
    \end{aligned}
    \label{eq:distill}
\end{equation}
where $p_T(\cdot\mid\cdot)$ and $p_{\boldsymbol{\theta}}(\cdot\mid\cdot)$ denote the teacher and student distributions, respectively. 
The superscript $(k)$ denotes the distribution over the top-$k$ tokens, giving rise to a sparse \ac{kl}.
For our experiments, we set $k = 256$ \citep[cf.][]{team2025gemma}.
The sparse \ac{kl} in Equation~\eqref{eq:distill} makes it possible to precompute and store teacher targets over the full distillation dataset.
As a result, the teacher does not need to be accessed directly during stage~II.
This is especially advantageous for long-context distillation, where querying an online teacher can become prohibitively costly. Scaling this regime efficiently will be an important focus of future work.

\textbf{Optional stage~III: expert merging.}
Stages~I--II can be applied either in a multi-task setting (one generalist student) or in a \emph{decentralized} setting where $K$ domain experts (e.g., math, code, STEM, etc.) are trained in parallel, all starting from the same initialized seed weights $\boldsymbol{\theta}^{(0)}$.
This branch-train-merge workflow mirrors a broader trend in post-training pipelines that target specific capabilities and later consolidate them into a single deployable model \citep{deepseekai2025deepseekv32,cohere2025commandA,team2026mimo-v2-flash}.
Concretely, after distilling linear experts $\{\boldsymbol{\theta}^{(i)}\}_{i=1}^K$, we form a single student via simple linear weight merging \citep{wortsman2022model}:
\begin{equation}
    \boldsymbol{\theta}_{\mathrm{merge}}=\sum_{i=1}^{K}\lambda_i\,\boldsymbol{\theta}^{(i)},
    \qquad \lambda_i\ge 0,\;\sum_{i=1}^{K}\lambda_i=1,
    \label{eq:linear-merge}
\end{equation}
with uniform weights by default and optional validation-tuned $\lambda_i$ when emphasizing particular capabilities.
In our setting, this enables \emph{capability patching}: researchers can independently improve a specific domain expert and update the final hybrid student by re-merging, without retraining the full model end-to-end.
Moreover, the expert-centric setup is particularly well-suited for applying domain-specific fine-tuning or on-policy distillation to each expert before merging, i.e., learning from self-generated trajectories with teacher feedback, which we leave for future work \citep{agarwal2024on-policy}.
For a brief overview of decentralized post-training pipelines and model merging, see Section~\ref{sec:related-work-model-merging}.

\section{Experiments}
\label{sec:experiments}

In this section, we apply our linearization protocol to both base models and instruction-tuned models from the Llama, Qwen, and Olmo families.
We conduct downstream evaluations of the resulting hybrid models on established benchmarks across two~important domains: \textbf{(1)} language understanding \& knowledge tasks, and \textbf{(2)} language generation \& reasoning tasks.
Across benchmarks, we compare our distilled xLSTM students both against its teacher model and 
state-of-the-art linearization alternatives, including {LoLCATs} \citep{zhang2025lolcats}, 
{RADLADS} \citep{goldstein2025radlads}, and Mamba-in-Llama \citep{wang2024mamba}.
We leverage \texttt{lm-eval} \citep{sutawika2025lm-eval} for conducting our evaluations (see Appendix~\ref{appendix:downstream-evals} for details).
For mathematical evaluations, we use the \texttt{Math-Verify} evaluation system.

\textbf{Metrics for effective distillation: teacher-recovery rate and tolerance-corrected win-and-tie rate.}
Similar to \citet{goldstein2025radlads}, we report the respective \emph{teacher-recovery rate} 
as a primary per-benchmark metric, defined as the ratio between student and teacher performance.
A recovery rate $>1$ indicates that the student exceeds its teacher on the respective benchmark.
We refer to Appendix Section \ref{appendix:downstream-evals} for absolute scores.
However, when comparing distilled models across a diverse suite of benchmarks, 
recovery rates alone do not quantify whether a student is a \emph{reliable} drop-in replacement.
In particular, simple aggregates of recovery (e.g., mean/median recovery) can obscure substantial 
regressions on a subset of tasks, and ratio-based summaries can be uninformative when the teacher
scores are small, yielding misleadingly large or noisy relative changes.
We therefore complement recovery rates with a \emph{tolerance-corrected win-and-tie} 
metric that summarizes \emph{task-level win-rate}
across benchmarks.
Following our definition of (approximately) lossless distillation (Appendix Section~\ref{sec:lossless-definitions}), 
we compute the win-and-tie rate $C_\alpha$, 
i.e., the fraction of benchmarks on which the student matches or 
exceeds teacher performance within a tolerance $\alpha$. 
This metric captures parity coverage across heterogeneous evaluations and 
distinguishes truly lossless distillation from partial recovery.
For compact model comparison, we report $\alpha^*$: the minimum tolerance $\alpha$ such that $C_\alpha \ge 0.5$.
Lower $\alpha^*$ indicates a better student, and thus a better distillation process, 
since less tolerance is required to match the teacher on half of the benchmarks.

\begin{figure*}
    \centering   
    \includegraphics[width=0.5\linewidth]{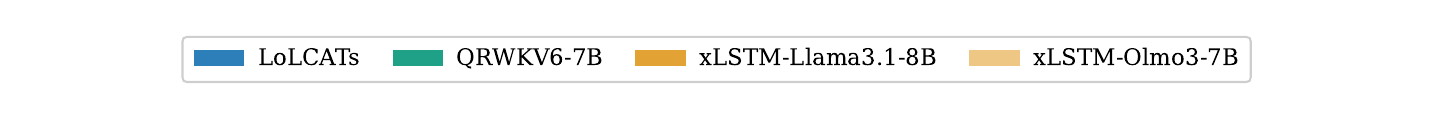}
    \begin{subfigure}{.5\linewidth}
        \includegraphics[width=\linewidth]{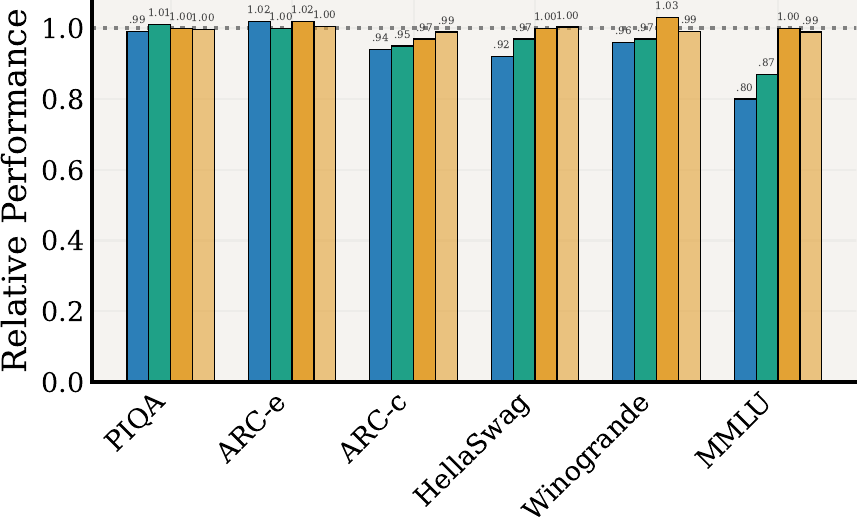}
        \caption{Language Understanding}
    \end{subfigure}%
    \hfill%
    \begin{subfigure}{.48\linewidth}
        \includegraphics[width=\linewidth]{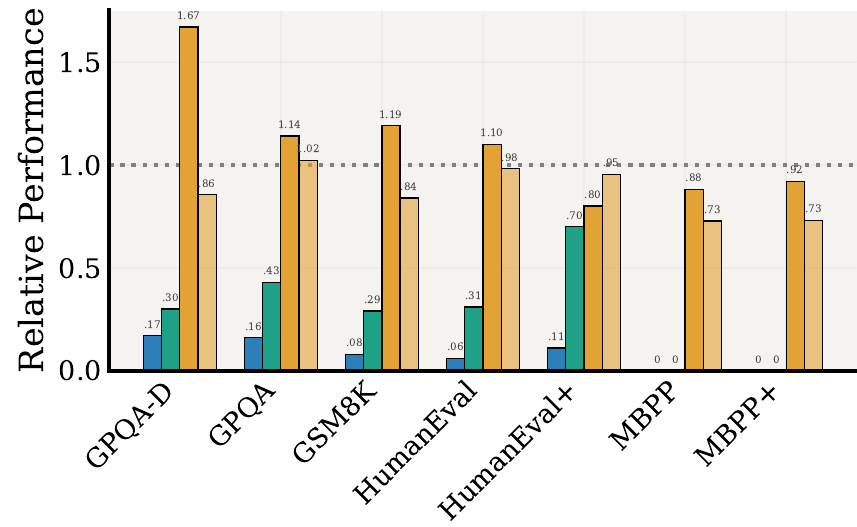}
        \caption{Language Generation}
    \end{subfigure}
    \caption{
        \textbf{Downstream evaluations} for \textbf{(a)} language understanding and \textbf{(b)} language generation tasks.
        We report the recovery rate relative to teacher scores for our mLSTM-based student and established baselines with comparable parameter counts. The dotted line at $1.0$ indicates parity with the Transformer teacher.
        Our model matches the teacher's performance across language understanding tasks, while exceeding the teacher on four of the considered generation tasks.
    }
    \label{fig:downstream-evals-base}
\end{figure*}

\subsection{Base Model Evaluation: Validating the Hybrid Architecture}
\label{sec:base-model-results}

To assess the generality of our linearization pipeline for base models, we distill both \llamaTeacher~and \olmoTeacher. 
Olmo's fully open pre-training corpus provides a unique opportunity 
to evaluate whether matching the teacher on the original data distribution improves distillation compared to using alternative public datasets.

\textbf{Experimental setup}.
For both models, we conduct stage~I hidden-state alignment over 655M~tokens with a sequence length of~4K using a standard linear-warmup to peak learning rate of~$10^{-2}$ and cosine decay to~$10^{-5}$. 
For Llama we leverage the Dolmino dataset\footnote{\url{https://huggingface.co/datasets/allenai/dolmino-mix-1124}}, and for Olmo we use the Dolmino~3 midtraining mix\footnote{\url{https://huggingface.co/datasets/allenai/dolma3_dolmino_mix-100B-1025}} released as part of \textsc{Olmo2} \cite{olmo2025olmo2} and \textsc{Olmo3} \cite{olmo2025olmo3} and maintain the originally proposed mixing weights.

For stage~II, we further distill our aligned Llama checkpoints on an additional 5~billion tokens from the same data mixes and context size as in phase I.
For Olmo, we extend the token budget to 20~billion tokens to align the budget with the protocol
used for instruction-tuned models (cf. Section \ref{sec:instruct-model-results}). 
For both models, we train using~$\gamma=0.9$ and~$\beta=0.1$ for \ac{ce} and \ac{kl} losses, respectively, and rewarm to a constant learning rate of~$10^{-5}$.
Moreover, we provide additional experiment details, including a description of training settings and hyperparameters in Appendix~\ref{appendix:exp-imp-details}.

\newpage

\textbf{Results}.
First, we evaluate our xLSTM-based students on six established multiple-choice (MC) and 
log-likelihood tasks, such as MMLU \citep{hendrycks2021measuring}, that test for general language understanding and knowledge.
Among publicly available baselines, \baselineLolcats~is distilled from the same \llamaTeacher~teacher, enabling a direct recovery-rate comparison, while \baselineQrwkvBase~is distilled from a Qwen-family teacher \citep{goldstein2025radlads}.
We observe that our distilled students achieve full (\llamaStudent) or near-full (\olmoStudent) teacher parity, while \baselineLolcats~and \baselineQrwkvBase~exhibit a significant performance gap. We report the respective teacher-recovery rates in Figure~\ref{fig:downstream-evals-base}a. Additionally, absolute scores are reported in Table \ref{tab:language-understanding}.
Next, we evaluate our distilled models on a broad battery of commonly used language generation and reasoning tasks that span important domains such as mathematics and coding \citep{cobbe2021training,austin2021program}. 
Unlike language understanding tasks, these benchmarks test the model's ability to produce consistent and relevant answers. In Figure~\ref{fig:downstream-evals-base}b, we report the recovery rate of our xLSTM-based students and established baselines (see Table \ref{tab:language-generation} for the raw scores).
We discover that prior methods exhibit significant performance gaps compared to the teacher, 
with \baselineLolcats~and \baselineQrwkvBase~both yielding $\alpha^\star=1.0$.
In contrast, our hybrid models achieve strong relative scores across most tasks, achieving $\alpha^\star=0.0$ for \llamaTeacher~and $\alpha^\star=0.01$ for \olmoTeacher.

\begin{figure*}
    \centering   
    \includegraphics[width=1.0\linewidth]{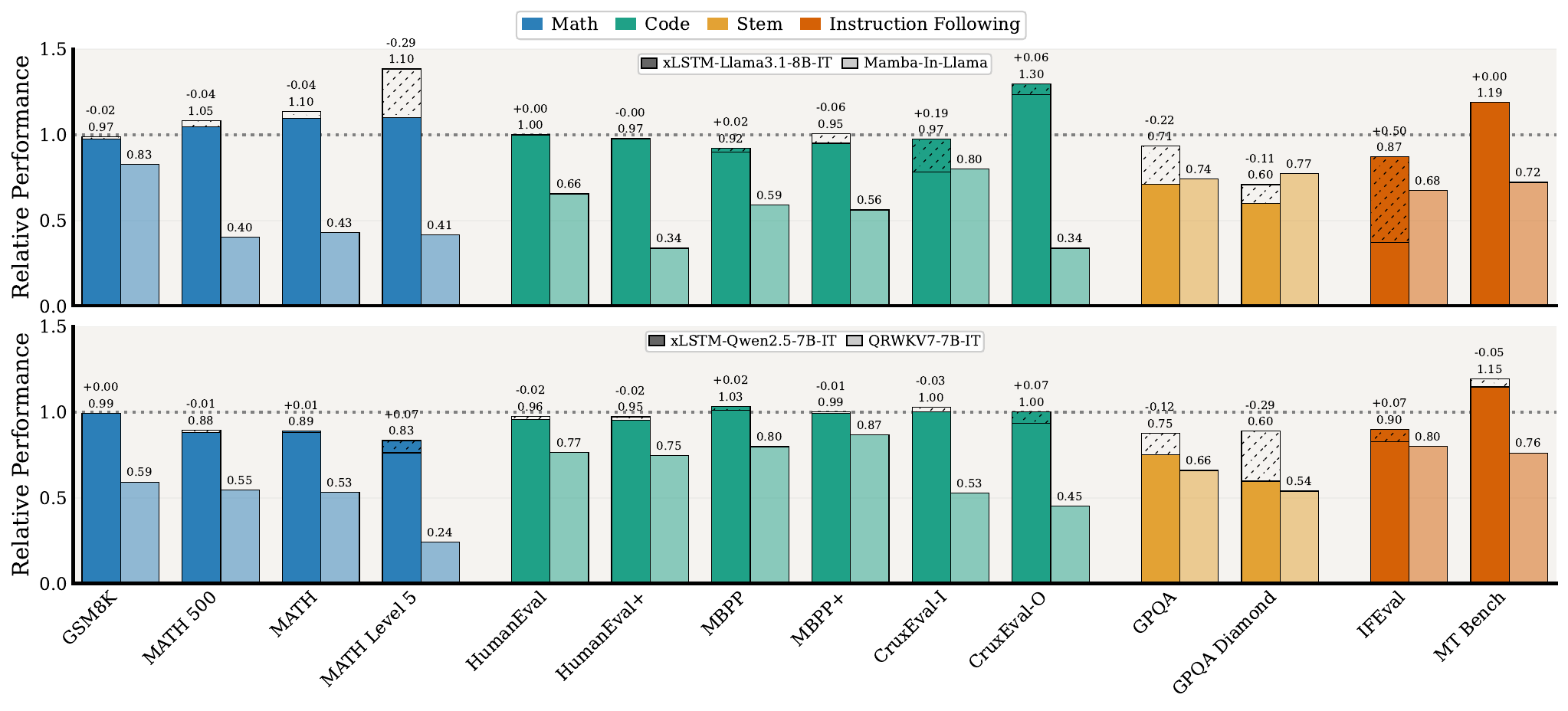}
    \caption{
        \textbf{Teacher-recovery rates for instruction-tuned xLSTM students and the effect of expert merging.}
        \textbf{Top:} \llamaStudentIT~distilled from \llamaTeacherIT~vs.\ \baselineMambaLlama;
        \textbf{Bottom:} \qwenStudentIT~distilled from 
        \qwenTeacherIT~vs.\ \baselineQrwkv.
        For each benchmark (x-axis; grouped by domain color), we report \emph{relative performance} as the student/teacher score ratio (y-axis); the dotted line at $1.0$ indicates parity with the Transformer teacher.
        For our method (left bar in each pair), the \emph{merged} student is shown. For a given task, the striped area on top of the bar indicates gains (colored) or losses (empty) compared to our linearized \textit{domain expert} before merging.
        For the baselines (right bar in each pair), the light bar shows the recovery rate.
    }
    \label{fig:downstream-evals-instruct}
\end{figure*}

\subsection{Instruction-Tuned Model Evaluation: Validating Decentralized Linearization}
\label{sec:instruct-model-results}

\textbf{Experimental setup}. Next, we apply our linearization pipeline to \emph{post-trained} models, focusing on \llamaTeacherIT~and \qwenTeacherIT. 
To ensure coverage of the capabilities of interest, we use our decentralized linearization scheme: starting from the teacher weights, we train four linearized specialists targeting \textbf{math}, \textbf{STEM}, \textbf{code}, and \textbf{instruction-following/chat}. 
Each expert is trained for $\sim$5B tokens on a domain-specific mixture constructed from Nemotron Nano-2/3 and \textsc{Olmo-3} data (\citealp{nvidia2025nemotron-nano2,nvidia2025nemotron3,olmo2025olmo3}; full mixtures and sampling weights in Table~\ref{tab:data-mixes}).
As for base models, we use the Stage~II objective in Eq.~\eqref{eq:distill} with $\gamma=0.9$ and $\beta=0.1$, and train with a constant learning rate of $7\times 10^{-7}$ for \textsc{Qwen} and $10^{-5}$ for \textsc{Llama}.
After training, we consolidate the specialists into a single deployable student via linear weight merging (Eq.~\eqref{eq:linear-merge}). We choose merge weights via lightweight sweeps on small sets of downstream evaluations and simple heuristics such as downweighting experts that underperform across domains.

\textbf{Results.}
Beyond language understanding benchmarks, we evaluate both the individual experts and the merged student on an expanded suite of generative benchmarks spanning our target domains (math, STEM reasoning, code, and instruction-following).
Figure~\ref{fig:downstream-evals-instruct} reports teacher-recovery rates for each benchmark and highlights the effect of merging by comparing the merged checkpoint with its constituent specialists.

We compare against instruction-tuned linearization baselines with aligned teachers.
For \llamaTeacherIT, we use \baselineMambaLlama, a Mamba2 attention hybrid that retains 50\% of the original softmax attention layers.
For \qwenTeacherIT, we compare to \baselineQrwkv~\citep{goldstein2025radlads}.
We find that our decentralized distillation pipeline yields strong linearized students that match their teachers on language-understanding benchmarks and code-generation evaluations, while recovering most of the teacher performance on mathematical reasoning tasks. We additionally assess instruction-following quality on MT-bench \citep{zheng2023judging} using LLM-as-a-judge, where GPT-5.1 grades generated responses. Across both the \textsc{Llama} and \textsc{Qwen} families, our students receive higher preference scores than their respective teachers (see Appendix \ref{appendix:add-results} Figure \ref{fig:mtbench} for detailed results). The largest remaining gap appears in STEM reasoning, where the merged student underperforms the dedicated STEM expert. 
Overall our merged students exhibit strong performance with \llamaStudentIT~and \qwenStudentIT~achieving $\alpha^\star=0.02$ and $\alpha^\star=0.05$ respectively.
We therefore conclude that expert training and subsequent model merging is a promising strategy to distill capabilities of interest in parallel and unify linearized models.

\textbf{Model merging discussion} 
Across both model families, we observe \emph{positive transfer} when unifying independently trained experts into a single hybrid student.
Most notably, merging substantially improves instruction-following for Llama, where we recover a large fraction of the gap on \textsc{IFEval} compared to the instruction expert alone.
At the same time, merging is not uniformly beneficial. Both students exhibit the most pronounced degradations on STEM-oriented evaluations (e.g., \textsc{GPQA} and \textsc{GPQA-Diamond}), indicating interference between domain updates.
In contrast, math and code capabilities are largely robust to merging, and for \qwenTeacherIT~we also observe comparatively minor changes in instruction-following.
Overall, these results validate that simple weight-space merging can be effective even for \emph{fully linearized} architectures, opening a practical path toward consolidating independently developed efficient students.

\subsection{Ablations}
\label{sec:ablations}

\textbf{Setup.} For the following ablations we distill linear student models from \llamaTeacher~with a token budget of ~2.5 billion tokens.

\textbf{Effect of mLSTM, SWA \& Sinks.}
Our hybrid approach utilizes a combination of mLSTM, \ac{swa} with a fixed size of 512 tokens and 4 sink tokens.
In Figure~\ref{fig:component_ablations}, we empirically validate the individual contributions of these components.
In addition, we compare against a pure linear attention baseline to understand the contribution of mLSTM.
Pure mLSTM exhibits a considerably lower loss than linear attention, which highlights the effectiveness of its gating mechanism.
The combination of mLSTM and \ac{swa} results in a striking improvement, pointing towards a harmonic relationship between modeling short and long-term dependencies via \ac{swa} and mLSTM, respectively.
We report additional analyses on the importance of sink tokens and \ac{swa} in Appendix~\ref{appendix-sec:architecture-ablation}. 
Moreover, we analyze the mixture weights produced by the data-dependent output gate and find that both contribute considerably to the final outputs (see Appendix~\ref{appendix:output-gate}). 

\textbf{Effect of distillation objective.}
The distillation objective is another critical component of every distillation recipe. 
Therefore, we conduct an ablation in which we vary the mixture weights $\gamma$ and $\beta$ for the CE and KL losses used in stage~II, respectively. 
We observe that~$\gamma=0.9$ and~$\beta=0.1$ result in a low cross-entropy loss, while preventing the student from drifting too far away from its teacher (see Appendix~\ref{appendix-sec:distill-objective}).
In contrast, training purely via KL resulted in worse performance compared to a mixed objective due to overconstraining the linear student to the teacher. This trend persists even when distilling on the teacher’s original training data, indicating that the gains are not merely an artifact of \textit{up-training} the student on higher-quality data.

\textbf{PEFT vs. FFT.}
Prior linearization recipes use \ac{peft} via low-rank adaptation (LoRA, \citealp{hu2022lora}) to recover performance lost during conversion \citep{zhang2025lolcats,nguyen2025lizard}. 
While LoRA is cheaper, it is also less expressive than \ac{fft}.
We find that \ac{fft} results in considerably improved performance and therefore adopt it by default in our distillation recipe (see Appendix~\ref{appendix-sec:peft-vs-fft}).

\begin{figure*}[!t]
    \centering
    \includegraphics[width=0.50\linewidth]{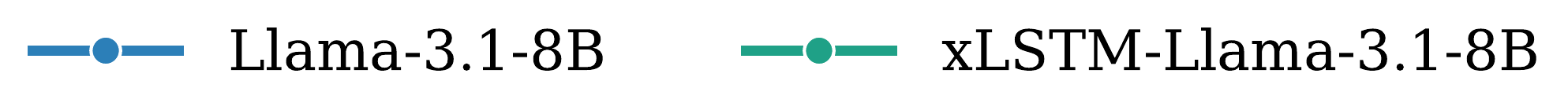}
    \begin{subfigure}{.33\linewidth}
        \includegraphics[width=\linewidth]{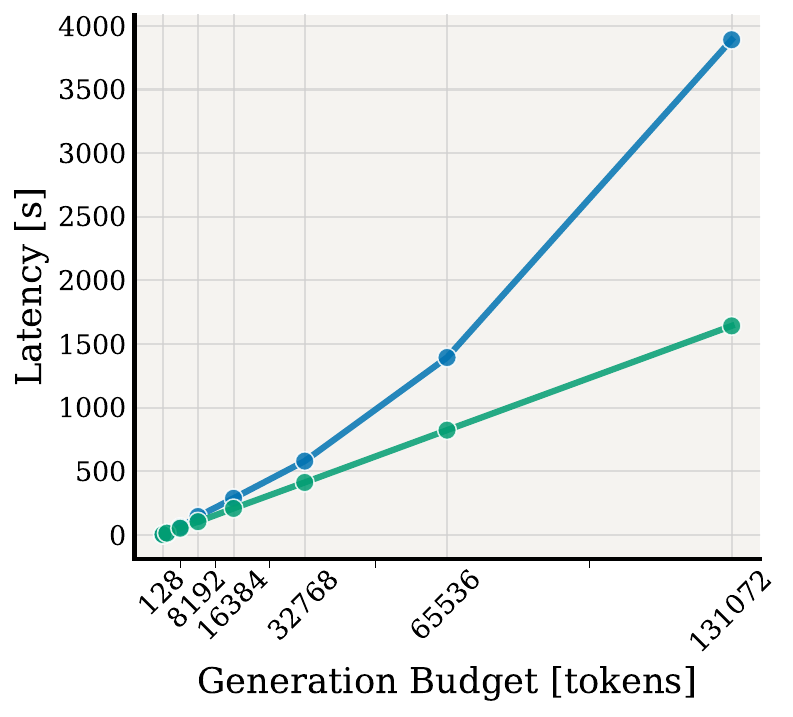}
        \caption{Latency, $B=1$}
    \end{subfigure}%
    \begin{subfigure}{.33\linewidth}
        \includegraphics[width=\linewidth]{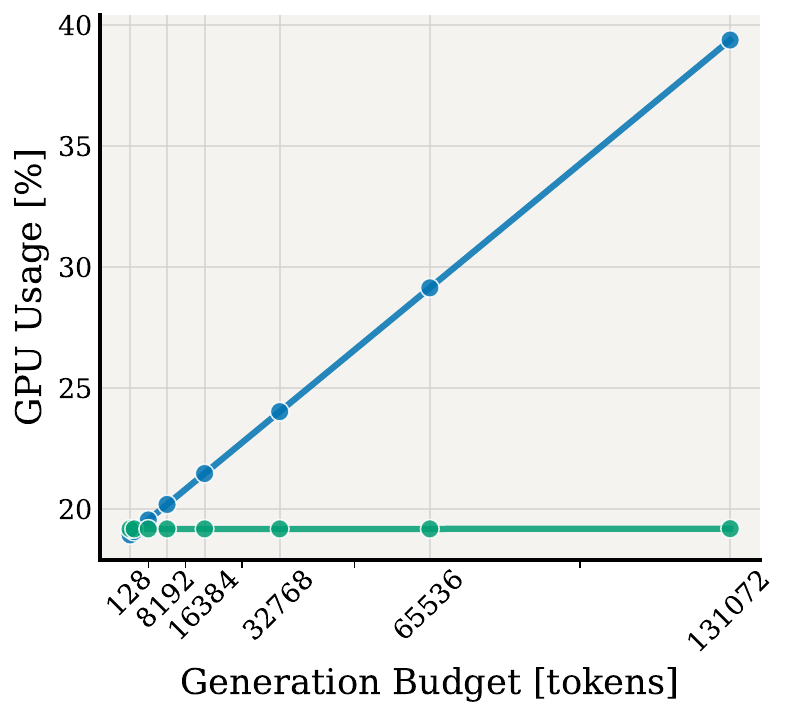}
        \caption{GPU RAM \%, $B=1$}
    \end{subfigure}%
    \begin{subfigure}{.33\linewidth}
        \includegraphics[width=\linewidth]{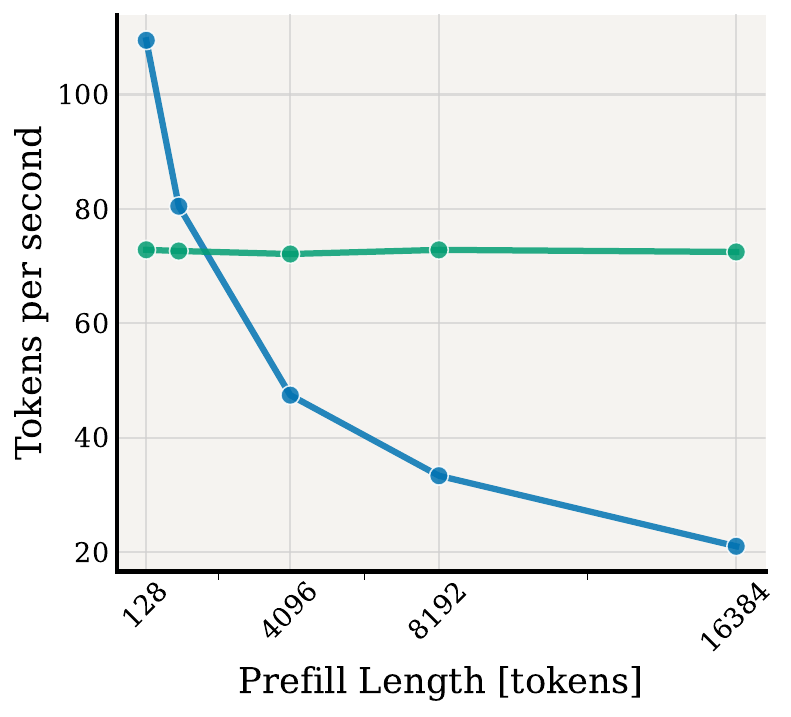}
        \caption{Throughput, $B=8$}
    \end{subfigure}
    \caption{
        Inference comparison for the \textbf{generation stage} between the Transformer-based teacher and our xLSTM-based student.
        In \textbf{(a)}, we show generation latency at different generation budgets ($B=1$).
        In \textbf{(b)}, we report the memory consumption in \% of GPU memory during the generation ($B=1)$.
        In \textbf{(c)}, we show the generation throughput when generating 100 tokens with varying prefill lengths and $B=8$.
    }
    \label{fig:inf-comparison-generation-main}
\end{figure*}

\subsection{Inference Comparison}
\label{sec:inference}

A key motivation for distilling Transformer teachers into recurrent xLSTM students is improved serving efficiency.
Following prior work \citep{de2024griffin,beck2025xlstm7b}, we report inference results separately for \emph{prefill} (prompt encoding) and \emph{generation} (autoregressive decoding) \citep{pope2023efficiently}.  Our inference tests are characterized by three core hyperparameters, the batch size~$B$, the context length~$C$, and the generation budget~$G$.
Both teacher and student are implemented in \texttt{transformers} \citep{wolf2020transformers} and optimized with \texttt{torch.compile}, FlashAttention \citep{dao2024flashattention2}, and fused mLSTM kernels \citep{beck2025tiled}; the student uses a static cache that stores mLSTM states plus sliding-window \ac{kv} (details in Appendix~\ref{appendix:inference}).

\textbf{Prefill.}
Figure~\ref{fig:inf-comparison-prefill} reports throughput and \ac{ttft} on a single H100~80GB for the largest feasible batch-size/context-length pairs.
Our hybrid student is consistently faster, with $\sim2\times$ higher throughput at $B{=}1,\,C{=}65$K and an overall $\sim2\times$ reduction in \ac{ttft}.

\textbf{Generation.}
Figure~\ref{fig:inf-comparison-generation-main} compares decoding latency, memory, and throughput.
Without prefill (to isolate decoding complexity), the student roughly halves latency and GPU memory at $G{=}131$K, while maintaining constant memory over time.
With prefill and $B{=}8$, the student achieves up to $\sim4\times$ higher generation throughput as context length grows; the teacher runs \ac{oom} at larger batches (Appendix~\ref{appendix:inference}).

\newpage

Overall, we demonstrate strong efficiency benefits of our xLSTM-based student at inference time, in terms of latency, throughput, and memory consumption.
We note that other linearized methods, such as {LoLCATs} \citep{zhang2025lolcats}, exhibit similar inference advantages (but with a larger student-teacher gap on downstream tasks) and therefore omit an additional comparison.

\textbf{Serving systems perspective.}
Beyond the experimental measurements reported here, production deployment of hybrid models requires integrating linear-complexity sequence-mixing layers into serving stacks such as vLLM or SGLang, which are built around paged \ac{kv}-memory abstractions such as PagedAttention \citep{kwon2023efficient}.
As hybrid and otherwise heterogeneous architectures become more common, including recent releases such as Qwen3.5 \citep{qwen3.5}, Nemotron~3 \citep{nvidia2025nemotron3}, and Olmo Hybrid \citep{merrill2026olmohybrid}, the system problem of efficiently scheduling, caching, and allocating memory across mixed layer types is becoming increasingly important.
Recent work such as Jenga makes this challenge explicit for heterogeneous \ac{llm} serving and points to the need for serving runtimes that can handle non-uniform memory and access patterns efficiently \citep{zhang2025jenga}.
We view this as essential systems work and as an important direction for research.

\section{Conclusion}
\label{sec:conclusion}
We presented a linearization pipeline that replaces quadratic softmax attention with an efficient mLSTM--\ac{swa} hybrid.
To assess whether distilled students are reliable drop-in replacements, we formalized lossless distillation through the \emph{Win-and-Tie rate} $C_\alpha$: the fraction of benchmarks on which the student matches or exceeds teacher performance (optionally within tolerance $\alpha$), and its critical tolerance $\alpha^*$, the minimum $\alpha$ such that $C_\alpha \ge 0.5$, i.e., the tolerance needed to match the teacher on at least half of benchmarks.
Across base and instruction-tuned teachers from the Llama, Qwen, and Olmo families, our xLSTM students attain substantially higher $C_\alpha$ and Pareto-dominate prior linearization methods across tolerances, reflecting more effective distillation across benchmarks rather than gains on a small subset.
As a result, our hybrid student models are prime candidates for a drop-in replacement of full-attention transformers when inference efficiency matters.
We further showed that \emph{distilling domain experts} and consolidating them via simple weight-space merging improves $C_\alpha$, indicating that weight merging remains effective after full linearization and enabling modular capability development and targeted updates.

\paragraph{Limitations and future work.}
The remaining deficits are most visible on synthetic long-context evaluations (e.g., Needle-in-a-Haystack) and on select reasoning benchmarks, where interference between independently trained experts can reduce recovery after merging. A key next step is to further probe which expert domains are most beneficial to distill in isolation and how to consolidate them more reliably. For long-context behavior, we plan to explore stronger attention hybrids and memory designs. Finally, we aim to scale this recipe to larger teachers, including Sparse-Mixture-of-Experts models, and to study on-policy distillation or RL-based expert refinement prior to merging.

\section*{Acknowledgments}
This work was supported by European Union’s Horizon Europe research and innovation programme under grant agreement number 101214398 (ELLIOT) and the Austrian Science Fund (FWF) 10.55776/COE12.
The ELLIS Unit Linz, the LIT AI Lab, the Institute for Machine Learning, are supported by the Federal State Upper Austria.
We acknowledge EuroHPC Joint Undertaking for awarding us access to Leonardo at CINECA, Italy, Deucalion at MACC, Portugal, and Discoverer at SofiaTech, Bulgaria.

\clearpage

\bibliography{references}
\bibliographystyle{plainnat}

\clearpage
\appendix
\startcontents[appendix]

\section*{Appendix}
\printcontents[appendix]{l}{1}{\setcounter{tocdepth}{2}}
\clearpage

\section{Definition of Knowledge Distillation Goals}
\label{sec:lossless-definitions}

\begin{nxaiinfo}[Lossless and $\alpha$-Tolerant Knowledge Distillation]
\textit{Knowledge distillation} (KD) aims to transfer the knowledge of a pre-trained \textit{teacher} model to a strictly more efficient \textit{student} model. KD is \textit{lossless} if the transfer process yields a student that matches or outperforms its teacher across a broad spectrum of downstream tasks under an identical evaluation protocol. \\

More formally, let $\mathcal B = \{b_1,\ldots,b_n\}$
be a set of benchmarks, $A_{S}(b)$ and 
$A_{T}(b)$ be the accuracies of the student
and the teacher on benchmark $b$, respectively.
A student is at least as good as the
teacher up to a tolerance of $\alpha \in [0,1]$:
\[A_S(b) \geq (1-\alpha) A_T(b)\]. 

\begin{align}
\mathbf{1}_\alpha(b) =
\begin{cases}
1 & \text{if } A_S(b) \ge (1 - \alpha)\, A_T(b), \\
0 & \text{otherwise}.
\end{cases}
\end{align}

The \emph{Win-and-Tie rate at a tolerance level of $\alpha$} 
of the student relative to the teacher is then
\begin{align}
C_\alpha = \frac{1}{|\mathcal{B}|} \sum_{b \in \mathcal{B}} \mathbf{1}_\alpha(b).
\end{align}

We can now define 
\emph{lossless distillation} when a student exhibits 
$C_0 \geq 0.5$, i.e. equal or better accuracy on 
at half of the benchmarks without any tolerance. 
\emph{$\alpha$-tolerant distillation} 
occurs if a student reaches a Win-and-Tie rate 
$C_\alpha \geq 0.5$ at tolerance level $\alpha$. \\

The quality of a distillation process can be assessed by 
\begin{align}
     \alpha^\star = \inf \{\alpha  \mid C_\alpha \geq 0.5\}
\end{align}
Smaller $\alpha^\star$ indicates a more conservative and
higher-quality distillation, as less tolerance is required
for the student to match the teacher on at least half of
the benchmarks.

\end{nxaiinfo}

\paragraph{Win-and-Tie rate curves.}
The definition above allows us to investigate 
the Win-and-Tie rate $C_\alpha$ in dependency 
of different values for $\alpha$ for each model. 
With increasing tolerance $\alpha$ the student is
more often considered as equal or better than the 
teacher, which means increasing $C_\alpha$. If the student
matches or outperforms the teacher on at least half of
the benchmarks, the models can be considered as at least 
equally performant and thus as a \emph{successful 
distillation process}. Win-and-tie rate curves 
show at which tolerance values, successful distillation 
would be reached. 

\paragraph{Model comparison and Pareto front.} 
These curves also allow for model comparison: the
higher the curve the better. The \textbf{Pareto-front} 
is at the top curve: at a given tolerance value, the
best pick is the method that provides the best
Win-and-Tie rate.

\begin{figure}
    \centering

    \begin{subfigure}[t]{0.42\linewidth}
    \centering
    \includegraphics[width=\linewidth]{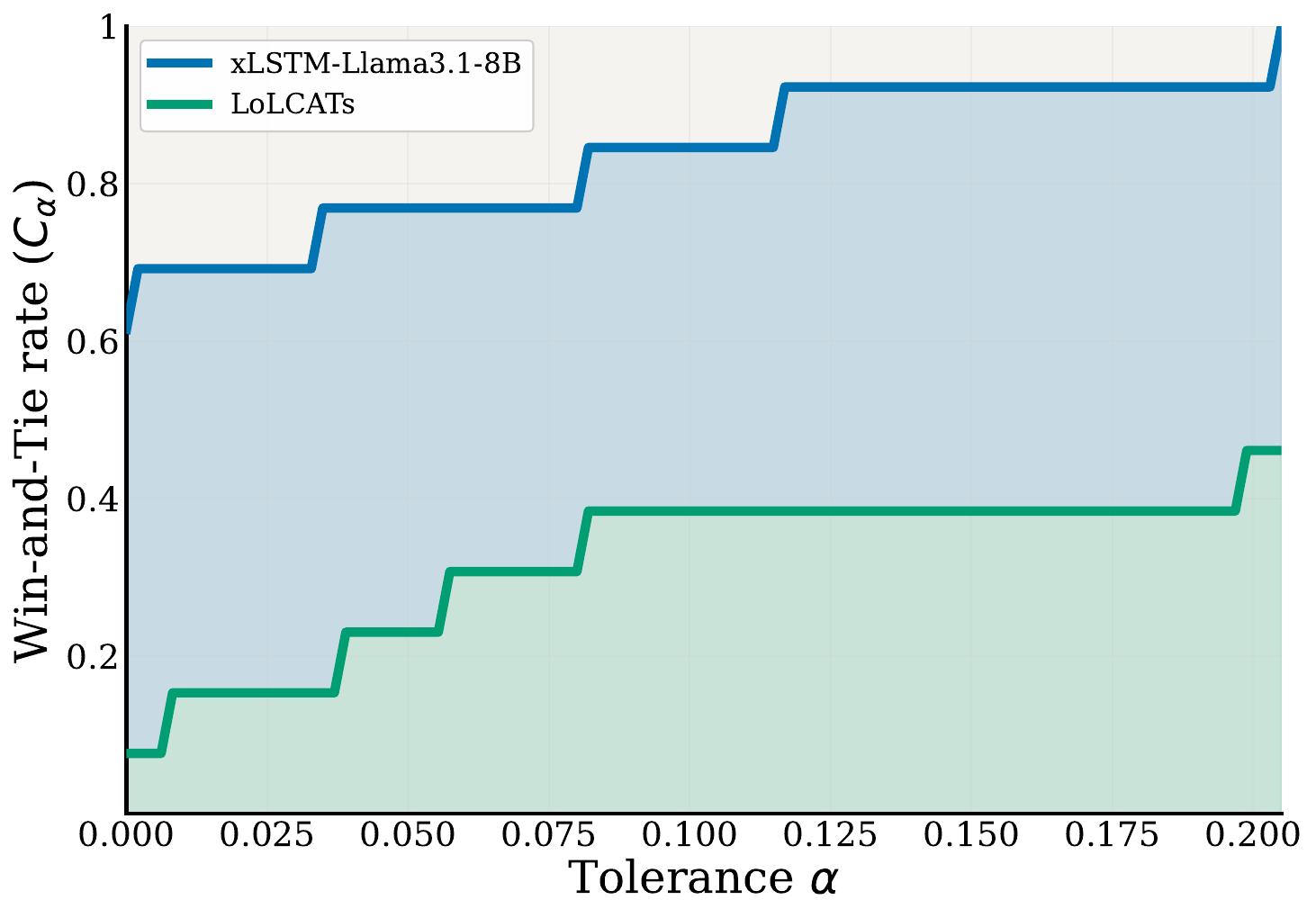}
    \label{fig:pareto-llama-base}
    \end{subfigure}
    \begin{subfigure}[t]{0.42\linewidth}
    \centering
    \includegraphics[width=\linewidth]{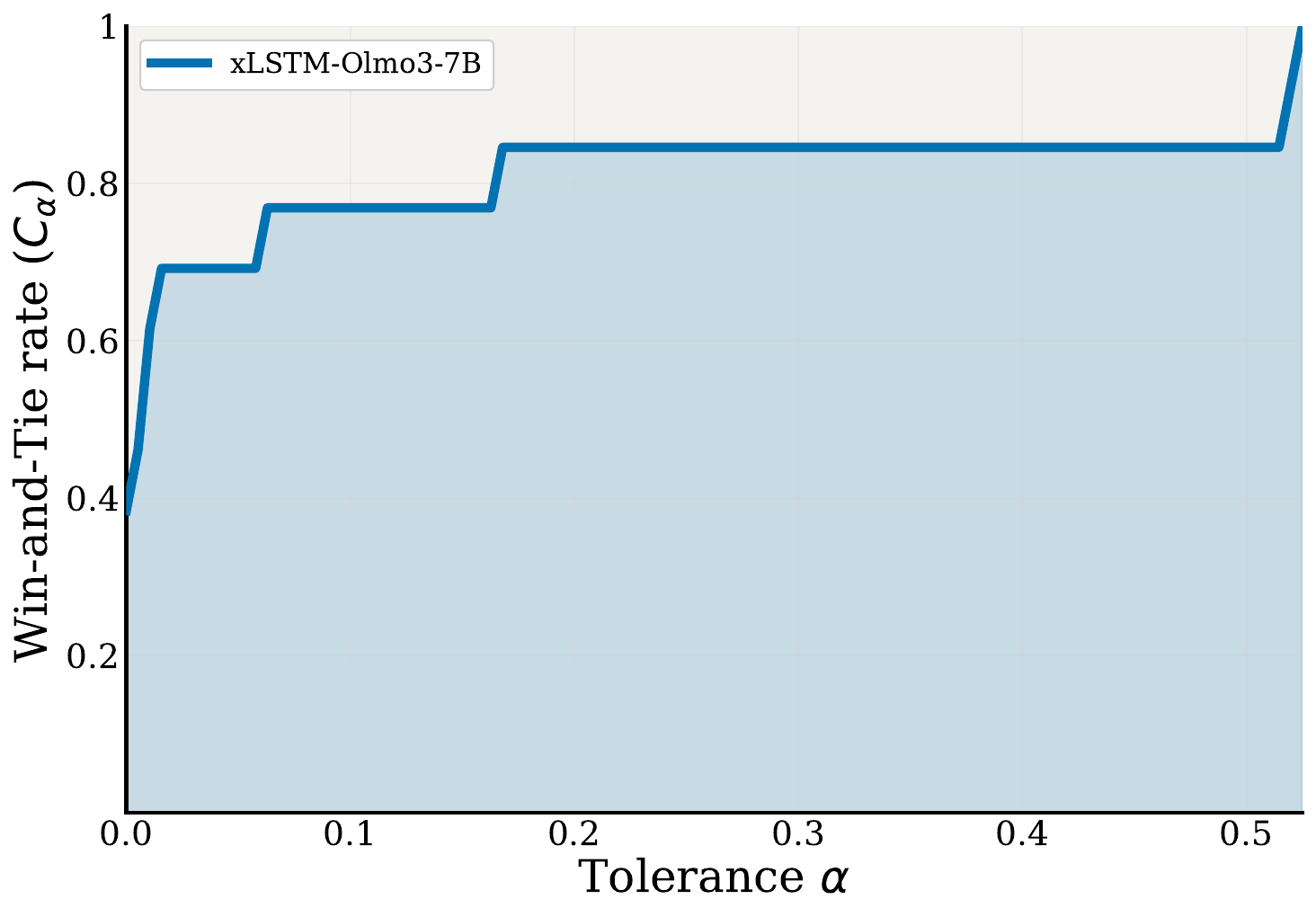}
    \label{fig:pareto-olmo-base}
    \end{subfigure}
    
    \vspace{-3mm} %
    
    \begin{subfigure}[t]{0.42\linewidth}
    \centering
    \includegraphics[width=\linewidth]{figures/pareto/win_tie_rate_llama_it_corrGK.pdf}
    \label{fig:pareto-llama-it}
    \end{subfigure}
    \begin{subfigure}[t]{0.42\linewidth}
    \centering
    \includegraphics[width=\linewidth]{figures/pareto/win_tie_rate_qwen_it_corrGK.pdf}
    \label{fig:pareto-qwen-it}
    \end{subfigure}

    \caption{Win-and-Tie rate ($C_\alpha$) curves of our xLSTM-distilled Models (\llamaStudent, \olmoStudent, \llamaStudentIT, \qwenStudentIT) and their respective best performing subquadratic baselines across generations benchmarks spanning math, code, STEM, and chat domains (for \olmoTeacher~there are no available Baselines at the time of writing). On the y-axis, we show the Win-and-Tie rate $C_\alpha$ between the student and teacher for a given tolerance $\alpha$. Distillation can be considered successful if the student matches or outperforms the teacher in 50\% of the benchmarks. Thus, $C_\alpha$ values above $0.5$ can be considered successful distillation attempts.}
    \label{fig:winandtie}
\end{figure}

\newpage

\section{Related Work}
\label{sec:related-work}

\subsection{Modern Recurrent and Hybrid Architectures.}
\label{sec:related-work-hybrid-archs}

\paragraph{Linear attention and \acs{ssm} alternatives to softmax attention.}
A broad line of work targets sub-quadratic sequence operators with linear-time training and constant-memory decoding.
Beyond \acp{ssm} such as Mamba \citep{dao2024transformers}, this space includes data-dependent convolutions such as Hyena \citep{poli2023hyena} and linear-attention families augmented with expressive gating mechanisms such as GLA, DeltaNet, Gated-DeltaNet, RWKV, and xLSTM \citep{yang2024gated,schlag2021deltanet,peng2023rwkv,beck2024xlstm}.
Among recurrent gated variants, xLSTM offers two operators.
\emph{sLSTM} uses a scalar state with exponential input gating.
\emph{mLSTM} maintains a matrix-valued state with head-wise scalar gates and is fully parallelizable, with a recurrent formulation for decoding.
These operators provide stable long-horizon memory, efficient rank-one fast-weight updates, and constant-size states at inference.
\vspace{-5pt}
\paragraph{Hybrid models: inter-layer vs.\ intra-layer.}
A growing literature blends quadratic attention with linear attention or \ac{ssm} primitives.
These works can be categorized into inter-layer and intra-layer hybrids.

\emph{Inter-layer hybrids} interleave attention and linear sequence-mixing blocks.
Notable early works include Jamba \citep{lenz2025jamba} and Zamba \citep{glorioso2024zamba}, which alternate between Mamba and global softmax attention layers.
Samba \citep{ren2025samba} replaces the global attention layers in earlier designs with sliding-window attention, yielding a fully linear architecture that combines global memory with precise short-range recall.
SWAX \citep{cabannes2025short} alternates sliding-window attention and mLSTM and applies stochastic window resizing to strengthen long-context capability.
Recent production-scale efforts extend this pattern.
{Nemotron-H} \citep{nvidia2025nemotron-h} and the {Nemotron-Nano-2} \citep{nvidia2025nemotron-nano2} replace most attention layers with Mamba.
{Qwen-Next} \citep{qwen2025qwen3-next} interleaves Gated DeltaNet layers and gated softmax attention layers.
{MiniMax-01} \citep{minimax2025minimax-01} combines Lightning Attention \citep{qin2024various}, a gated linear-attention variant, with global softmax attention layers.
Recently, {GPT-OSS} combines global attention with sliding window attention \citep{openai2025gpt-oss}.

\emph{Intra-layer hybrids} fuse attention and linear or SSM branches within a block and mix outputs through addition or learned gates. Head-wise designs allocate some heads to attention and others to \acp{ssm}.
Sequence-wise designs split tokens by absolute or relative position.
Recent synchronous designs allow linear mixers and attention layers to process the full input sequence at the same time.
Representative head-wise models include Hymba \citep{dong2025hymba}, which assigns half of the heads to parallel attention and the remainder to Mamba.
Sequence-wise hybrids include TransMamba \citep{li2026transmamba}, which can switch between attention and \ac{ssm} mechanisms at different transition points for different layers.
As an early influential \ac{swa} plus linear hybrid, BASED \citep{arora2024simple} utilized linear attention to compress tokens outside the sliding window into a compact global state. 

Our method fits the intra-layer category.
Every layer combines sliding-window attention with mLSTM.
Unlike Hymba, we do not separate sliding-window and linear branches across heads.
Each head models both local and global dependencies with both branches.
We chose this design for two reasons.
In distillation, the teacher's attention heads are already pre-trained, and assigning local or global roles would require analysis.
The design also grants each head more expressive power.
Unlike BASED, we do not restrict the linear branch to tokens that leave the sliding window.
Both branches model the full input sequence and can exploit their complementarity.
This design has recently also been shown to be effective for pre-training at smaller scales \citep{irie2025blending}.
We fuse the outputs by repurposing the mLSTM output gate, which yields a data-dependent combination of the two attention streams.
In distillation, we also find that modeling sink tokens is important.
Similar interventions appear in work on KV-cache compression \citep{xiao2024efficient} and in hybrids such as Hymba.
Unlike Hymba, which introduces learned meta tokens, we simply include the first four sink tokens in the sliding window.
Recent work also seeks to mitigate the emergence of attention sinks through pre-training interventions \citep{openai2025gpt-oss, miller2023attention}.
For those models, there is no need to handle sinks explicitly.

\subsection{Linearizing LLMs.}
\label{sec:related-work-linearization}

To reduce the prohibitive inference cost of Transformers on long sequences, recent work studies \textit{linearization}, a post-training procedure that replaces some or all softmax self-attention layers in a pre-trained model with linear-time sequence mixers such as gated RNNs or state-space models.
Notable examples include T2R \citep{kasai2021finetuning}, SUPRA \citep{mercat2024supra}, LoLCATs \citep{zhang2025lolcats}, Mamba-in-Llama \citep{wang2024mamba}, RADLADS \citep{goldstein2025radlads}, MOHAWK \citep{bick2024transformers} and Llamba \citep{bick2025llamba}.
Most methods copy compatible weights from the teacher into the student, which yields far greater data efficiency than training from scratch.
During the initial alignment stage, LoLCATs, MOHAWK, and Llamba match hidden states with an MSE loss, and MOHAWK and Llamba also match attention maps, which improves alignment but requires materializing the attention matrix with $O(n^2)$ memory.
We follow the recipe of copying weights and then calibrating a small set of new gating and head-projection parameters with MSE.

Adaptation strategies divide into low-rank updates and full fine-tuning.
LoLCATs and Lizard utilize low-rank adapters to reduce training cost.
Mamba-in-Llama, Llamba, MOHAWK, and RADLADS perform full-model updates, which are more compute-intensive but reduce approximation error between student and teacher.
Supervision choices also differ.
LoLCATs and Lizard optimize next-token cross-entropy, while Mamba-in-Llama, Llamba, MOHAWK, and RADLADS add logit alignment with a KL loss term.
We adopt a mixed objective that combines cross-entropy loss with a \ac{kl} penalty, and we align only $k$ sampled logits so that teacher outputs can be precomputed once, similar to \cite{team2025gemma}.

Architecturally, some systems hybridize linear sequence operators with variants of softmax attention, and others are fully softmax-free.
LoLCATs and Lizard propose intra-layer hybrids that mix sliding-window attention with a linear path, with Lizard adding gated linear attention and global meta tokens.
Mamba-in-Llama and MOHAWK study inter-layer hybrids in which only some layers are linearized \citep{bick2024transformers}.
RADLADS and Llamba convert all attention to linear-time mixers.
We follow the intra-layer route, pairing sliding-window attention with an mLSTM path and adding sink tokens, then gating the two paths for data-dependent mixing.

Training budgets and datasets vary widely, from about~20M to 12B~tokens.
RADLADS relies on DCLM \citep{li2024datacomp} and adds OpenThoughts \citep{guha2025openthoughts} for hybrid reasoning models, MOHAWK uses C4 \citep{raffel2020exploring}, and Llamba reports gains from FineWeb-Edu \citep{penedo2024fineweb}.
Lizard and LoLCATs distill on small instruction datasets, for example, Alpaca.
Our schedule uses about 650M~tokens for hidden-state matching and 1.3B~tokens for end-to-end fine-tuning on a Dolmino-derived mid-training mixture.

\subsection{Decentralized model training and weight merging.}
\label{sec:related-work-model-merging}

A growing set of post-training pipelines decentralize (often within a research organization) by training multiple capability- or domain-specialized variants and then consolidating them into a single deployable model.
Early on, Branch-Train-Merge (BTM) showed that independently trained domain experts can be ensembled or collapsed back into a single model via parameter averaging \citep{li2022branch-train-merge}.
Recent large-scale systems follow related patterns: DeepSeek-V3.2 unifies expert behaviors by sampling from specialists and then performing SFT on the resulting traces to consolidate capabilities \citep{deepseekai2025deepseekv32}, while Command-A reports merging specialists in a way closely aligned with our proposed workflow \citep{cohere2025commandA}.
Beyond direct weight fusion, MiMo Flash V2 uses multi-teacher on-policy distillation as a mechanism to combine models through teacher feedback on self-generated trajectories \citep{team2026mimo-v2-flash}.
When consolidation is performed in weight space, simple averaging can already be surprisingly effective (e.g., weight soups) \citep{wortsman2022model}, but more robust methods such as TIES explicitly mitigate parameter interference by resolving sign conflicts and trimming small-magnitude updates \citep{yadav2023ties-merging}.

\section{Extended Background}

\subsection{Receptive Field of SWA}
\label{appendix:receptive-field-of-swa}

Although a depth-$L$ stack of \ac{swa} layers has a nominal receptive field of $LW$, in practice the effective receptive field grows much more slowly and is biased toward recent tokens.
Empirical measurements, as well as signal-propagation arguments, suggest sublinear growth in $L$ \citep{xiao2025why}.
This mirrors classic results for deep convolutional networks, where the effective receptive field is Gaussian-like and occupies only a small fraction of the theoretical context \citep{luo2016understanding}.

\subsection{Attention Sinks}
\label{appendix:attn-sinks-perspective}

Transformers often place large, persistent attention mass on initial tokens (e.g., \texttt{<BOS>}). 
\citet{barbero2025why} argue that sinks are a useful stabilizer that prevents over-mixing through depth and preserves token identity, explaining why $\emph{sink patterns}$ emerge broadly even when those tokens are semantically irrelevant.
Sink behavior has been previously exploited for effective KV cache compression.
StreamingLLM \citep{xiao2024efficient} preserves a small sink prefix (1 -- 4 tokens) plus a sliding window for recent context and evicts the remaining tokens in the cache.
This compression partially recovers full-attention quality, improves length-generalization, and yields substantial decoding speedups.
A complementary perspective is that row-wise softmax in attention forces every head to allocate its entire probability mass across the sequence, encouraging spurious focus on sinks. 
Minimal fixes have been proposed, replacing $\mathrm{softmax}(x)_i=\tfrac{e^{x_i}}{\sum_j e^{x_j}}$ with $\mathrm{softmax}(x)_i=\tfrac{e^{x_i}}{e^{b}+\sum_j e^{x_j}}$, where $b$ is either set to 0 or a learned bias \citep{miller2023attention, openai2025gpt-oss}. 
This is effectively equivalent to adding a no-op \emph{null} key and value.

\section{Experimental \& Implementation Details}
\label{appendix:exp-imp-details}

All experiments were run on 8~H100 GPUs using PyTorch \ac{fsdp}. 
We configured our training with a global batch size of~64 (using gradient accumulation), mixed precision (bfloat16 for operations, float32 for gradient all-gather), and gradient clipping at a threshold of~1.0 for full finetuning.
To maximize GPU utilization, input sequences were packed to fill the maximum context length. 
We found that preserving the attention mask across these packed sequences, rather than truncating it, improved performance for our hybrid architecture. 
This finding is consistent with prior work \citep{buitrago2025understanding}.

\subsection{Hyperparameters and Data Mixes}
\label{appendix:exp-data}

\textbf{Hyperparameters.} Table~\ref{tab:hyperparams} summarizes the hyperparameters used throughout our distillation pipeline. We instantiate the student with an xLSTM $[0{:}1]$ configuration (mLSTM-only), using 32 mLSTM heads with head dimension 128 and rotary position embeddings, and train at a context length of 4096. Optimization is performed with AdamW (weight decay 0.1) at a global batch size of 64. We enable sequence packing to maximize context utilization while preserving attention masks across packed sequences, which improves throughput without truncating examples. In \textbf{Phase I}, we perform layer-wise hidden-state alignment with an MSE objective (cosine learning-rate schedule with peak LR $10^{-2}$), keeping the optimization focused on the newly introduced mixer/gating parameters. In \textbf{Phase II}, we switch to end-to-end knowledge distillation with a warmup followed by a low constant learning rate ($10^{-5}$ for \textsc{LLaMa}, $7\times 10^{-7}$ for \textsc{Qwen}). The Phase II stage interpolates between next-token cross-entropy and KL distillation, using weights $\gamma=0.9$ (\textit{CE}) and $\beta=0.1$ (\textit{KL}).

\begin{table}
    \centering
    \caption{Hyperparameters.}
    \footnotesize\begin{tabular}{ll}
\toprule
\textbf{Setting} & \textbf{Value} \\
\midrule
mLSTM/sLSTM configuration & \texttt{[1:0]} \\
mLSTM head dimension & 128 \\
Number of mLSTM heads & 32 \\
Position embeddings & true \\
Context size & 4096 \\
Total token budget & 20B (5B per expert) \\
Weight decay & 0.1 \\
Optimizer & AdamW \\
Batch size & 64 \\
Sequence packing & true \\
Learning rate (Phase I: MSE matching) & Cosine schedule, max LR $1\mathrm{e}{-2}$ \\
Learning rate (Phase II: Knowledge Distillation) & Warmup + LR ($1\mathrm{e}{-5}$ Llama, $7\mathrm{e}{-7}$ Qwen) \\
CE loss weight & 0.9 \\
KL loss weight & 0.1 \\
Linear merge weights (Llama) & $\texttt{math}\cdot 0.35 + \texttt{code}\cdot 0.35 + \texttt{stem}\cdot 0.20 + \texttt{chat}\cdot 0.10$ \\
Linear merge weights (Qwen) & $\texttt{math}\cdot 0.20 + \texttt{code}\cdot 0.40 + \texttt{stem}\cdot 0.15 + \texttt{chat}\cdot 0.25$ \\
\bottomrule
\end{tabular}

    \label{tab:hyperparams}
\end{table}

\textbf{Data Mix.} Table~\ref{tab:data-mixes} outlines the domain-specific data mixtures used to train the individual experts(\textit{math}, \textit{code}, \textit{stem}, \textit{chat/instruction-following}), as well a the multi-task mixture used for a \textit{generalist}-\textsc{xlstm} student. Across both \textsc{LLama} and \textsc{Qwen} variants, the data mixes are mostly constructed from \textit{Nemotron Nano Pre- and Post-training} datasets. Moreover, for some of the experts (e.g., code), we enhance the datamix with appealing domain-targeted data, such as \textit{OpenCodeInstruct} and \textit{Dolci-Think-SFT}. The stem expert differs the most between the two model families; the \textsc{Llama} stem expert was trained solely on the \textit{Nemotron STEM Post-Training} dataset, whereas for \textsc{Qwen} we enhanced our datamix with more specialized \textit{STEM/MCQ} traces, to minimize the gap to the teacher.

\begin{table}
    \centering
    \small
    \setlength{\tabcolsep}{4pt}      %
    \renewcommand{\arraystretch}{0.95} %
    \caption{Data mixes for expert and multi-task linearization.}
    \footnotesize\begin{tabular}{lllr}
\toprule
\textbf{Model} & \textbf{Dataset} & \textbf{Split} & \textbf{Mixing Weight (\%)} \\
\midrule

\multicolumn{4}{l}{\textit{xLSTM-Llama Math Expert}} \\
\midrule
 & nvidia/Nemotron-Pretraining-SFT-v1          & math   & 50   \\
 & nvidia/Llama-Nemotron-Post-Training-Dataset & math   & 25   \\
 & nvidia/Nemotron-Post-Training-Dataset-v2    & math   & 25   \\
\midrule

\multicolumn{4}{l}{\textit{xLSTM-Llama Code Expert}} \\
\midrule
 & nvidia/Llama-Nemotron-Post-Training-Dataset & code   & 21   \\
 & nvidia/Nemotron-Post-Training-Dataset-v2    & code   & 1    \\
 & nvidia/Nemotron-Pretraining-SFT-v1          & code   & 64   \\
 & nvidia/OpenCodeInstruct                     &        & 11   \\
 & allenai/Dolci-Think-SFT-Python              &        & 3    \\
\midrule

\multicolumn{4}{l}{\textit{xLSTM-Llama STEM Expert}} \\
\midrule
 & nvidia/Nemotron-Post-Training-Dataset-v1    & stem   & 100  \\
\midrule

\multicolumn{4}{l}{\textit{xLSTM-Llama STEM Expert FT}} \\
\midrule
 & nvidia/Nemotron-Post-Training-Dataset-v2    & stem   & 100  \\
\midrule

\multicolumn{4}{l}{\textit{xLSTM-Llama Instruction Following Expert}} \\
\midrule
 & nvidia/Nemotron-Pretraining-SFT-v1          & general & 90  \\
 & nvidia/Nemotron-Post-Training-Dataset-v1    & chat    & 10  \\
\midrule

\multicolumn{4}{l}{\textit{xLSTM-Llama Instruction Following Expert FT}} \\
\midrule
 & nvidia/Nemotron-Post-Training-Dataset-v2    & chat    & 100 \\
\midrule

\multicolumn{4}{l}{\textit{xLSTM-Qwen Math Expert}} \\
\midrule
 & nvidia/Nemotron-Pretraining-SFT-v1          & math   & 76   \\
 & nvidia/Llama-Nemotron-Post-Training-Dataset & math   & 20   \\
 & nvidia/Nemotron-Post-Training-Dataset-v2    & math   & 4    \\
\midrule

\multicolumn{4}{l}{\textit{xLSTM-Qwen Code Expert}} \\
\midrule
 & nvidia/Llama-Nemotron-Post-Training-Dataset & code   & 20   \\
 & nvidia/Nemotron-Post-Training-Dataset-v2    & code   & 1.2  \\
 & nvidia/Nemotron-Pretraining-SFT-v1          & code   & 60   \\
 & nvidia/OpenCodeInstruct                     &        & 12   \\
 & allenai/Dolci-Think-SFT-Python              &        & 6.8  \\
\midrule

\multicolumn{4}{l}{\textit{xLSTM-Qwen STEM Expert}} \\
\midrule
 & nvidia/Nemotron-Pretraining-Specialized-v1  &        & 90.6 \\
 & nvidia/Nemotron-Science-v1                  & MCQ    & 3.4  \\
 & nvidia/Nemotron-Post-Training-Dataset-v2    & stem   & 6    \\
\midrule

\multicolumn{4}{l}{\textit{xLSTM-Qwen Instruction Following Expert}} \\
\midrule
 & nvidia/Nemotron-Pretraining-SFT-v1          & general & 88  \\
 & nvidia/Nemotron-Post-Training-Dataset-v2    & chat    & 8   \\
 & nvidia/Nemotron-Instruction-Following-Chat-v1 & chat\_if & 4 \\
\midrule

\multicolumn{4}{l}{\textit{Multi-task}} \\
\midrule
 & nvidia/Nemotron-Pretraining-SFT-v1          & math    & 19    \\
 & nvidia/Llama-Nemotron-Post-Training-Dataset & math    & 5     \\
 & nvidia/Nemotron-Post-Training-Dataset-v2    & math    & 1     \\
 & nvidia/Nemotron-Pretraining-Specialized-v1  & stem    & 22.66 \\
 & nvidia/Nemotron-Science-v1                  & MCQ     & 0.84  \\
 & nvidia/Nemotron-Post-Training-Dataset-v2    & stem    & 1.5   \\
 & nvidia/Nemotron-Pretraining-SFT-v1          & general & 22    \\
 & nvidia/Nemotron-Post-Training-Dataset-v2    & chat    & 2     \\
 & nvidia/Nemotron-Instruction-Following-Chat-v1 & chat\_if & 1   \\
 & nvidia/Llama-Nemotron-Post-Training-Dataset & code    & 5     \\
 & nvidia/Nemotron-Post-Training-Dataset-v2    & code    & 0.3   \\
 & nvidia/Nemotron-Pretraining-SFT-v1          & code    & 15    \\
 & nvidia/OpenCodeInstruct                     &         & 3     \\
 & allenai/Dolci-Think-SFT-Python              &         & 1.7   \\
\bottomrule
\end{tabular}

    \label{tab:data-mixes}
\end{table}

\subsection{Generalist vs. Merged Expert Student}
\label{sec:generalistvsexpert}

To assess the effectiveness of decentralized expert training, we compare a \emph{generalist} student trained on the multi-task mixture against a \emph{merged expert} student obtained by training four domain specialists and combining them via linear merging. In the expert setting, each specialist is trained on its corresponding domain mixture (Table~\ref{tab:data-mixes}) for $\sim$5B tokens, for a total of 20B tokens, and then merged using the fixed weights in Table~\ref{tab:hyperparams}. In the generalist setting, a single student is trained for the same 20B token budget on the multi-task mixture, keeping architecture and optimization matched.

Table~7 shows that the \emph{generalist} student trained on the multi-task mixture lags behind the \emph{merged-expert} student, with the largest gaps appearing on specialized evaluations that directly probe individual capabilities. \llamaStudentIT~ consistently outperforms \llamaStudentITGeneralist~ on math and reasoning (MATH500 $0.54$ vs.\ $0.37$, MATH $0.55$ vs.\ $0.37$), code generation (HumanEval $0.63$ vs.\ $0.50$, HumanEval+ $0.56$ vs.\ $0.47$), and instruction-following (IFEval $0.69$ vs.\ $0.49$, MT-Bench $6.05$ vs.\ $5.45$). \qwenStudentIT~ similarly remains ahead of \qwenStudentITGeneralist~ across the same specialized benchmarks (e.g., MATH $0.66$ vs.\ $0.62$, MATH Level~5 $0.42$ vs.\ $0.36$, GPQA-D $0.26$ vs.\ $0.19$, MT-Bench $5.96$ vs.\ $5.55$).

We attribute this behavior to the multi-task distillation. Although the total token budget for distillation is the same, the generalist allocates fewer updates to the specialized traces that matter for our benchmarks (math, code, and instruction-format data), whereas each expert is trained on a single-domain data distribution. Linear merging then retains much of this specialization in the merged checkpoint, in line with decentralized post-training recipes~\citep{cohere2025commandA}.

\section{Additional Results}
\label{appendix:add-results}

In this section, we provide additional details on experiments and provide additional results to complement the main text.

\begin{figure}
  \centering
  \includegraphics[width=\linewidth]{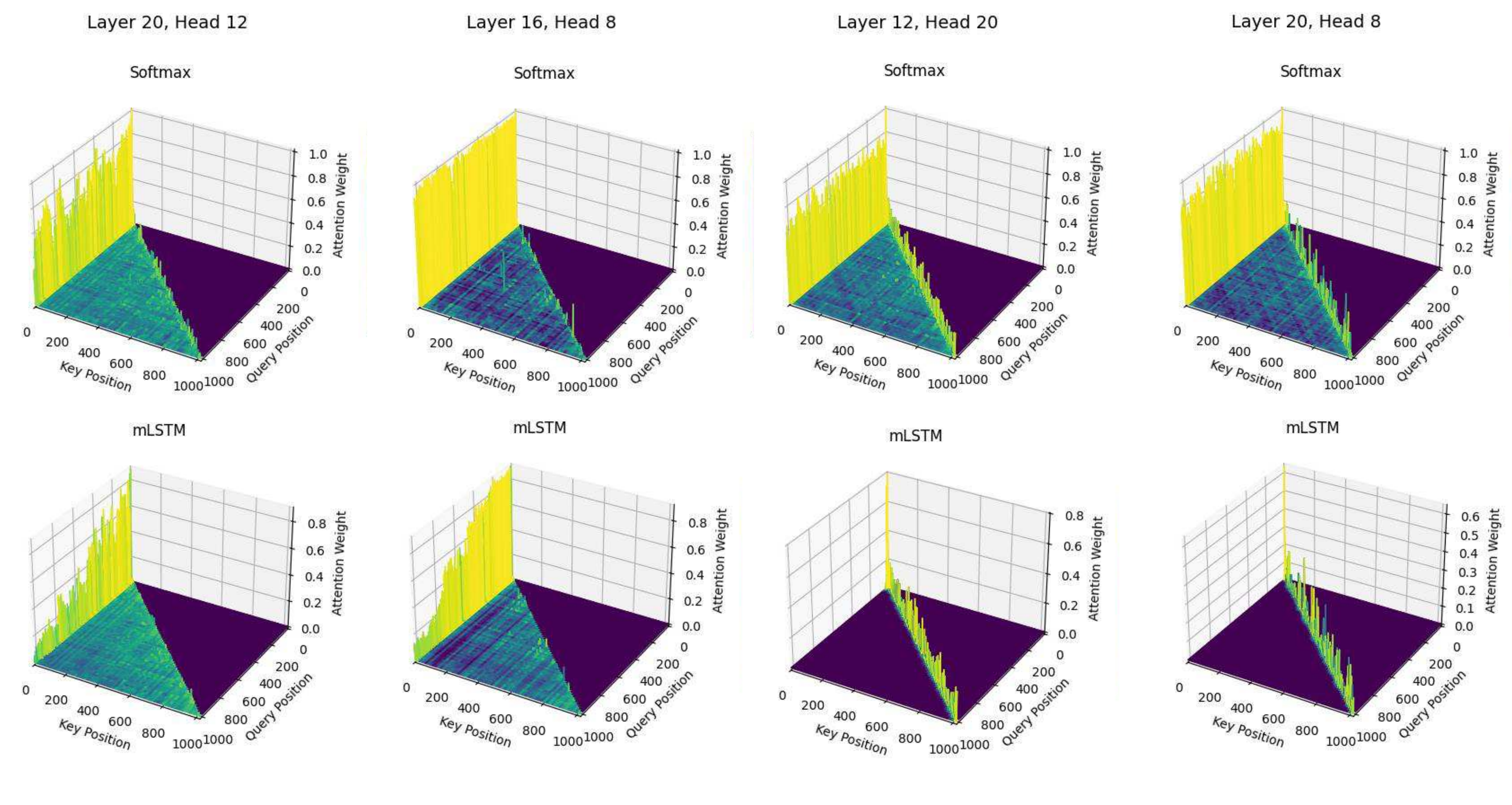}
  \caption{Illustration of \textbf{attention sinks} in the Llama~3.1~8B teacher model.}
  \label{fig:sink-patterns}
\end{figure}

\subsection{Sink Analysis}

As described in Section~\ref{sec:background}, Transformers often place large, persistent attention mass on the first on a small set of initial tokens.  
\citet{barbero2025why} argue that sinks are a useful stabilizer that prevents over-mixing through depth and preserves token identity. 
Consequently, many established LLMs exhibit sink patterns if they are not equipped with counter-measurements to prevent them from emerging. 
To illustrate the sink patterns in our Llama~3.1~8B teacher model, in Figure~\ref{fig:sink-patterns} we plot attention maps for two layers and two heads in a 3-dimensional grid.
Indeed, we find that the Transformer teacher puts the majority of its attention mass on the first token.
In contrast, when analyzing the distilled mLSTM-only checkpoints, we find that they struggle to represent the sink patterns present in the pre-trained teacher model. We observe that this forgetting effect worsens as the input sequence length grows. We suspect this is attributable to the decaying effect of the mLSTM's forget gate.

Therefore, modeling sink behavior is critical for strong performance.
In our experiments, we empirically confirmed that modeling sink patterns in combination with the sliding window results in a considerably lower loss and stronger downstream performance, as illustrated in our components ablation in Figure~\ref{fig:component_ablations}.

\subsection{Output Gate Analysis}
\label{appendix:output-gate}

As described in Section~\ref{sec:method} combine the individual outputs of the mLSTM and \ac{swa} components via a data-dependent output gate. 
To better understand the relative contributions of each component to the final prediction of our student models, we analyze the activations of the output gate.
To this end, we forward 128~sequences (sequence length~4096) through our model and record the average activation values for each layer and attention head, resulting in a layer $\times$ head gating matrix.

\begin{figure}
  \centering

  \begin{subfigure}[t]{0.48\linewidth}
    \centering
    \includegraphics[width=\linewidth]{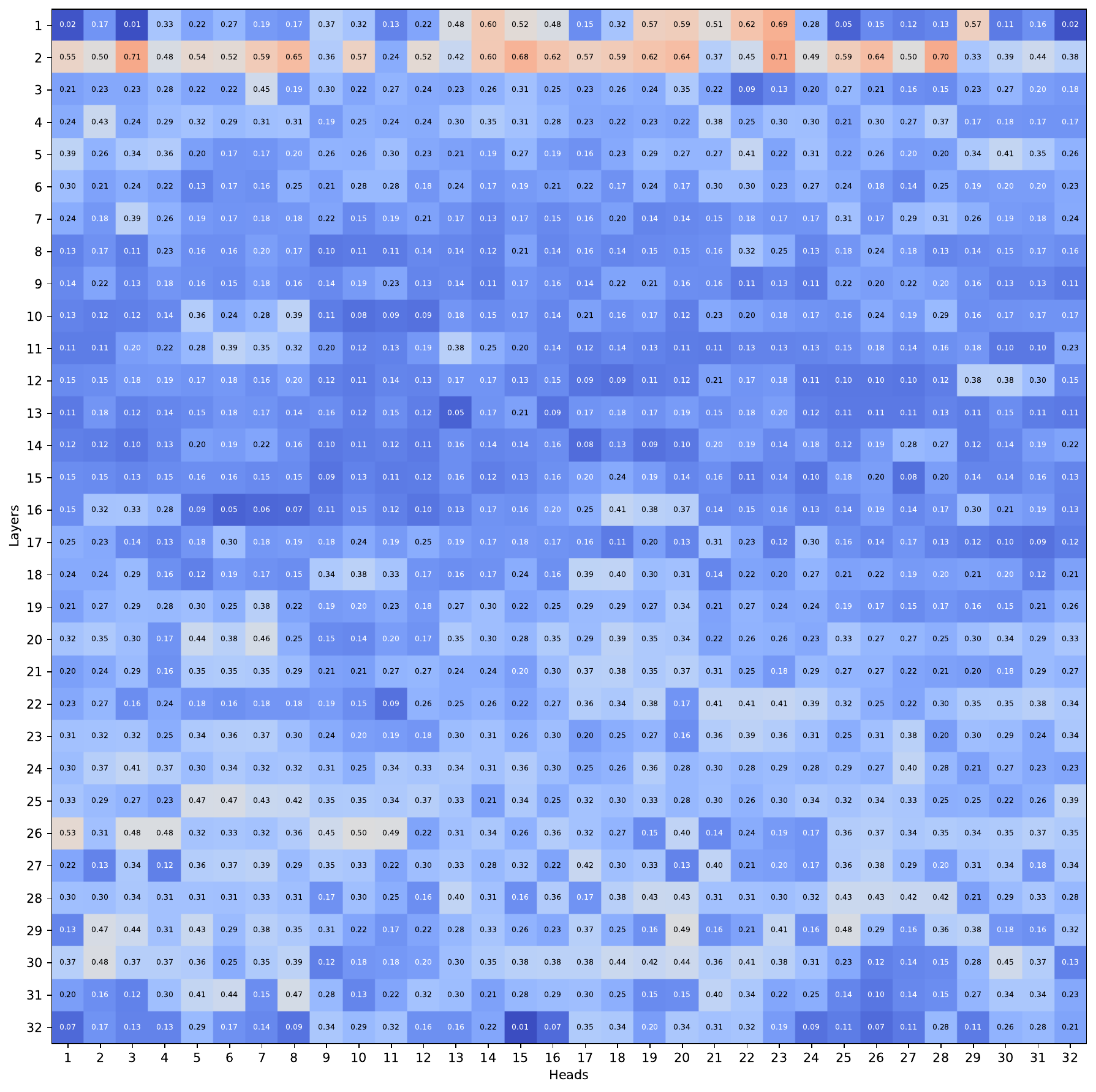}
    \caption{\llamaStudent}
    \label{fig:ogate-llama-base}
  \end{subfigure}
  \hfill
  \begin{subfigure}[t]{0.48\linewidth}
    \centering
    \includegraphics[width=\linewidth]{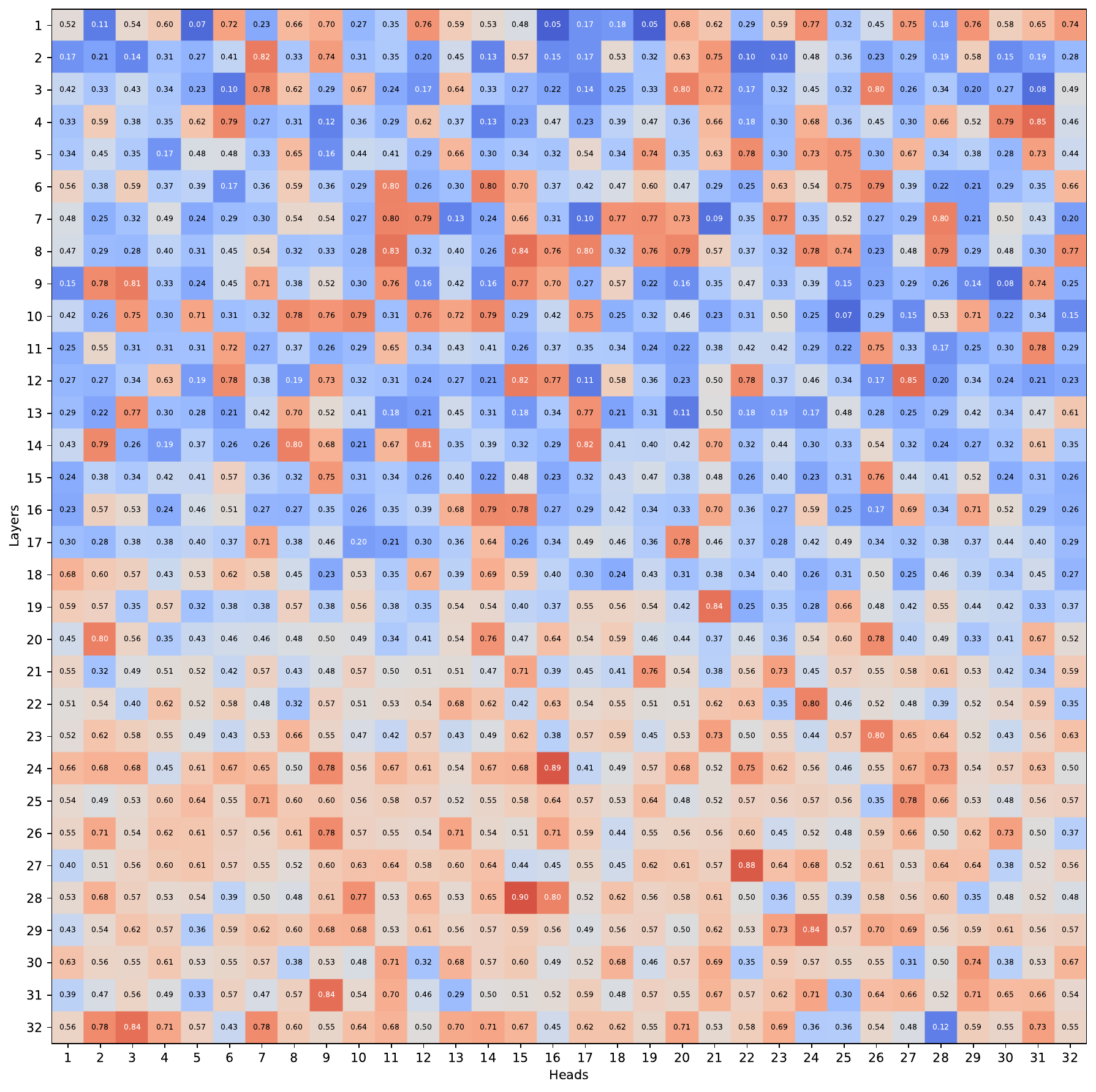}
    \caption{\olmoStudent}
    \label{fig:ogate-olmo-base}
  \end{subfigure}

  \vspace{0.6em} %

  \begin{subfigure}[t]{0.48\linewidth}
    \centering
    \includegraphics[width=\linewidth]{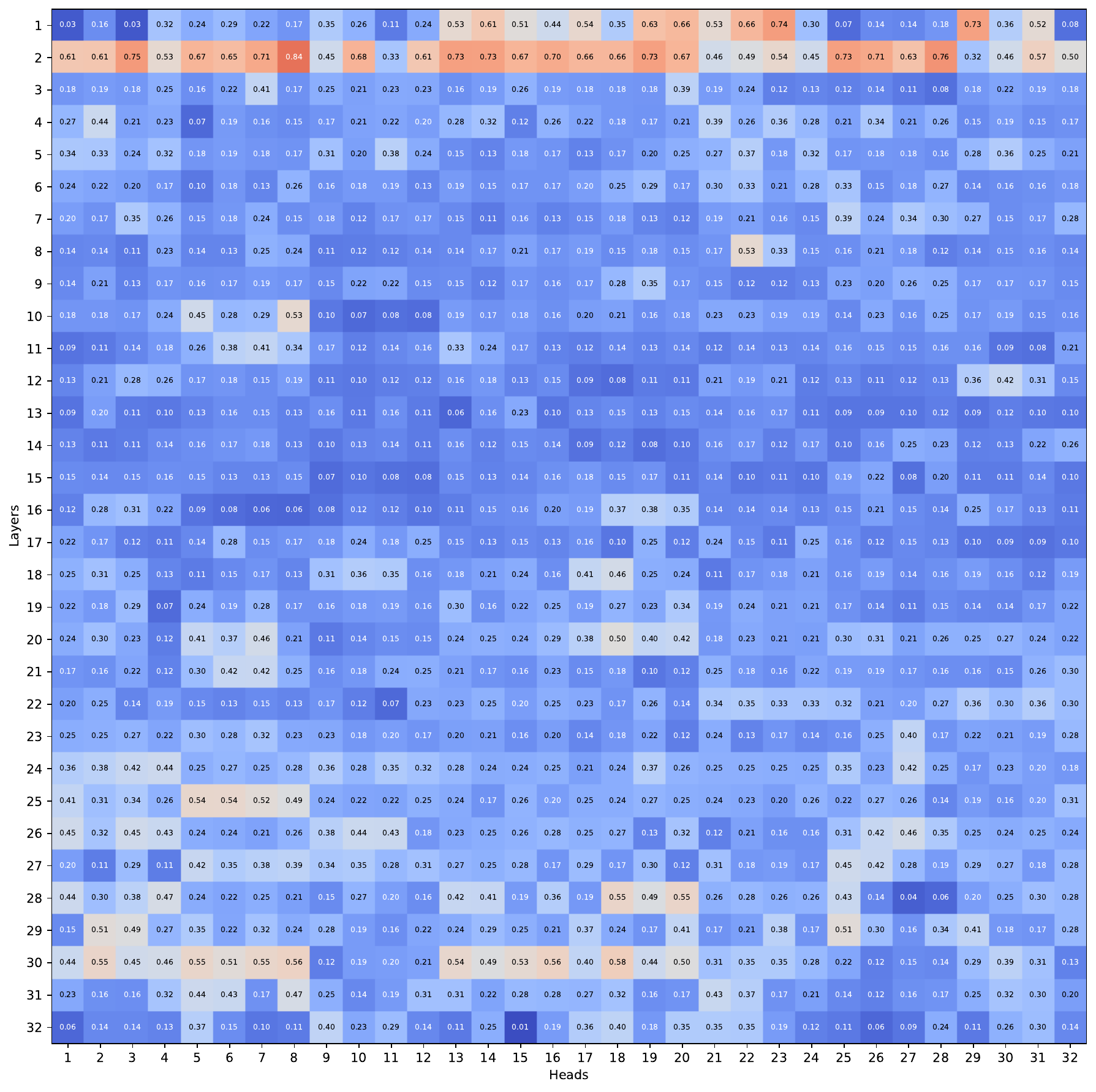}
    \caption{\llamaStudentIT}
    \label{fig:ogate-llama-it}
  \end{subfigure}
  \hfill
  \begin{subfigure}[t]{0.48\linewidth}
    \centering
    \includegraphics[width=\linewidth]{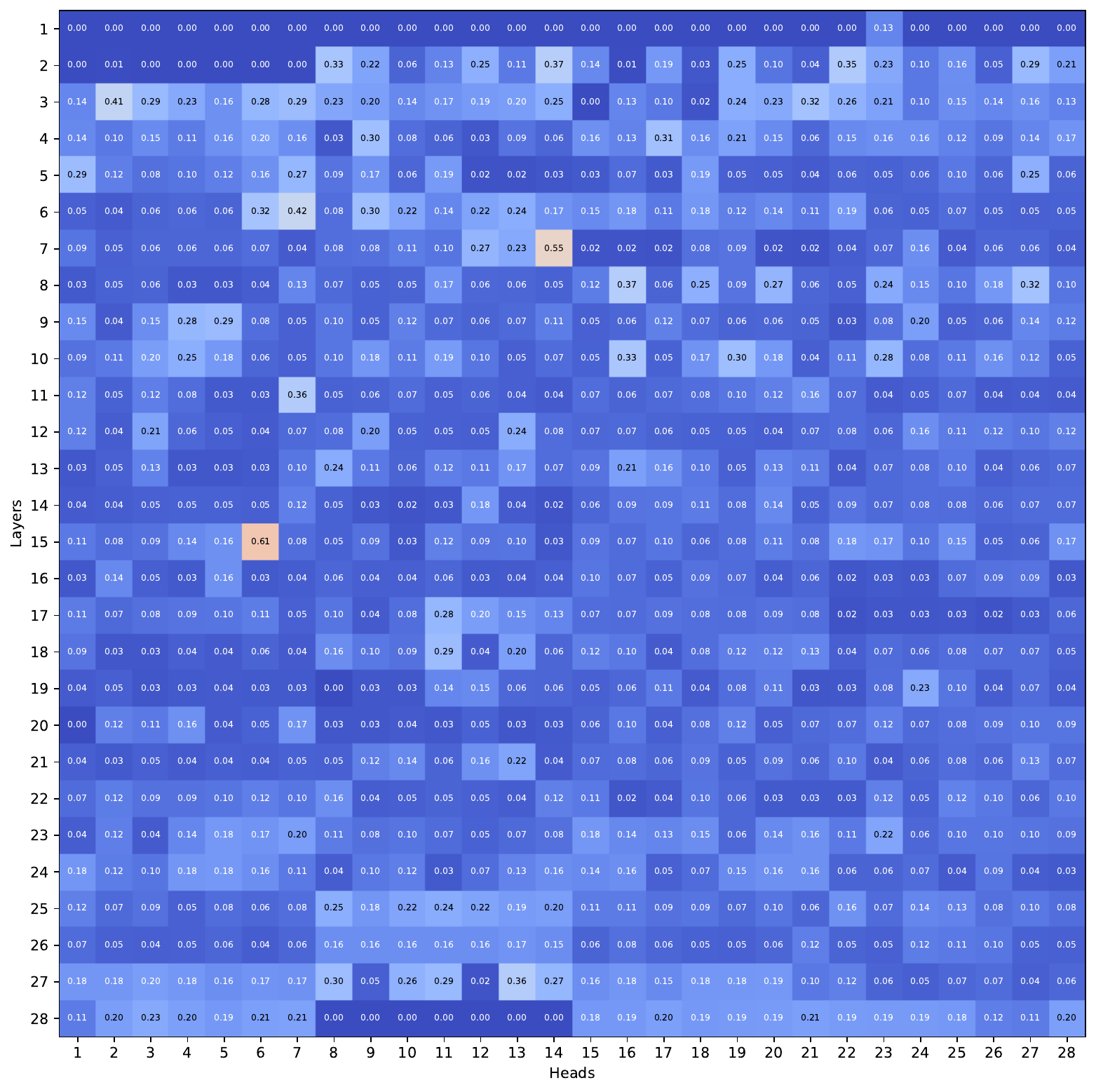}
    \caption{\qwenStudentIT}
    \label{fig:ogate-qwen-it}
  \end{subfigure}

  \vspace{-0.3cm}

  \begin{subfigure}[t]{\linewidth}
    \centering
    \includegraphics[width=\linewidth]{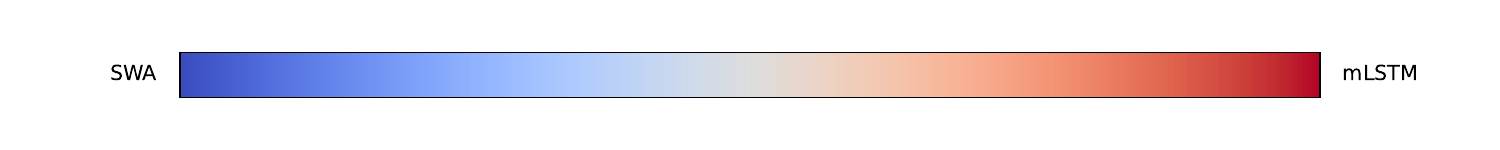}
    \label{fig:ogate-colorbar}
  \end{subfigure}

    \vspace{-0.8cm}
  
  \caption{
    Illustration of the data-dependent \textbf{output gate} output matrix.
    After training, we analyze how the mLSTM and \ac{swa} contributions are mixed across layers and heads. The plots above show the median output gate's sigmoid activation across layers and heads over 128 randomly drawn data samples with a context size of 4096. For a given example, 0 (blue) indicates that only sliding window is used, whereas an activation of 1 (red) means only mLSTM is used. We observe balanced to high mLSTM activations across model families. 
  }
  \label{fig:output-gate-grid}
\end{figure}

In Figure \ref{fig:output-gate-grid}, we observe that both components contribute significantly to the final output prediction across all layers.
Noticeably, mLSTM dominates in the first two layers, suggesting that the global contextual information carried by mLSTM blocks is integrated early on in the layer stack.
The middle layers (3 -- 16) are predominantly influenced by \ac{swa}, while the final layers (17 -- 32) exhibit a more balanced contribution from both components, with neither clearly dominating.

\subsection{Downstream Evaluations}
\label{appendix:downstream-evals}

After training, we evaluate our distilled student models on downstream tasks, which we group into 4~categories:
\begin{itemize}
    \item \textbf{Language Understanding}: log-likelihood and commonly used multiple-choice benchmarks
    \item \textbf{Language Generation \& Reasoning}: mathematics, coding and other established reasoning benchmarks
    \item \textbf{Language Generation Quality via MT-Bench}: across 6~tasks contained in MT-Bench \citep{zheng2023judging}
    \item \textbf{Needle in a Haystack}: we evaluate long context retrieval via Needle in a Haystack tasks.  
\end{itemize}

We conduct all our evaluations using \texttt{lm-eval} released by \citet{sutawika2025lm-eval}.
To ensure a consistent and fair comparison, we maintain the same number of few-shot examples, context lengths, and generation budgets across all teacher and student models.
Where available, we adopt the same evaluation settings as used by \citet{lambert2024tulu} and \citet{touvron2023llama}.
We provide all evaluation configurations for \texttt{base} and \texttt{instruct} models in Tables~\ref{tab:eval-configs} and~\ref{tab:eval-configs-instruct}, respectively.
\begin{table}
    \centering
    \caption{Downstream \textbf{evaluation configurations} for language understanding, generation, and quality benchmarks for all student and teacher \texttt{base} models.}
    \footnotesize\begin{tabular}{lrr}
\toprule
\textbf{Task} & \textbf{\# of shots} & \textbf{Generation Budget} \\
\midrule
\multicolumn{3}{l}{\textit{Language Understanding}} \\
\midrule
PIQA & 0 & -- \\
ARC-e & 0 & -- \\
ARC-c & 25 & 100 \\
HellaSwag & 10 & -- \\
Winogrande & 5 & -- \\
MMLU & 5 & 10 \\
\midrule
\multicolumn{3}{l}{\textit{Language Generation}} \\
\midrule
GPQA Diamond CoT & 0 & 2048 \\
GPQA Main CoT & 0 & 2048 \\
GSM8K & 8 & 1024 \\
HumanEval & 0 & 1024 \\
HumanEval Plus & 0 & 1024 \\
MBPP & 3 & 256 \\
MBPP Plus & 3 & 1024 \\
\midrule
\multicolumn{3}{l}{\textit{Language Quality}} \\
\midrule
MT-Bench & -- & 1024 \\
\bottomrule
\end{tabular}

    \label{tab:eval-configs}
\end{table}
\begin{table}
    \centering
    \caption{Downstream \textbf{evaluation configurations} for language understanding, generation, and quality benchmarks for all student and teacher \texttt{instruct} models.}
    \footnotesize\begin{tabular}{lrrcc}
\toprule
\textbf{Task} & \textbf{\# of shots} & \textbf{Gen Budget} & \textbf{Chat} & \textbf{Multiturn ICL} \\
\midrule
\multicolumn{5}{l}{\textit{Language Understanding}} \\
\midrule
PIQA & 0 & -- & \cmark & \xmark \\
ARC-e & 0 & -- & \cmark & \xmark \\
ARC-c & 25 & 100 & \cmark & \cmark \\
HellaSwag & 10 & -- & \cmark & \cmark \\
Winogrande & 5 & -- & \cmark & \cmark \\
MMLU & 5 & 10 & \cmark & \cmark \\
\midrule
\multicolumn{5}{l}{\textit{Language Generation}} \\
\midrule
GPQA Diamond CoT & 0 & 2048 & \cmark & \xmark \\
GPQA Main CoT & 0 & 2048 & \cmark & \xmark \\
GSM8K CoT & 5 & 3072 & \cmark & \cmark \\
MATH 500 & 0 & 3072 & \cmark & \cmark \\
MATH & 0 & 3072 & \cmark & \cmark \\
Math Level 5 & 4 & 1024 & \cmark & \cmark \\
HumanEval (64) Instruct & 0 & 3072 & \cmark & \xmark \\
HumanEval+ (64) Instruct & 0 & 3072 & \cmark & \xmark \\
MBPP & 3 & 1024 & \cmark & \cmark \\
MBPP+ & 3 & 1024 & \cmark & \cmark \\
Cruxeval-O & 0 & 3072 & \cmark & \cmark \\
Cruxeval-I & 2 & 3072 & \cmark & \cmark \\
\midrule
\multicolumn{5}{l}{\textit{Language Quality}} \\
\midrule
MT-Bench & -- & 1024 & \cmark & \cmark \\
\bottomrule
\end{tabular}

    \label{tab:eval-configs-instruct}
\end{table}

\textbf{Language Understanding \& Knowledge.}
To complement the relative teacher scores reported in Figure~\ref{fig:downstream-evals-base}, we report the raw performance scores across all language understanding tasks for all student and teacher models in Table~\ref{tab:language-understanding}. 
We find that our distilled student models match and in some tasks even slightly exceed their respective teacher performances. 
In contrast, other linearization recipes fall short of their respective teachers, exhibiting a significant performance gap. 
\begin{table}
    \centering
    \caption{
        Raw scores for \textbf{downstream evaluations} on \textbf{language understanding} tasks.
        Our models perform comparably to their \llamaTeacher, \olmoTeacher, \llamaTeacherIT, and \qwenTeacherIT~teacher models across language understanding tasks.
    }
    \label{table:qa_tasks}
    \setlength{\tabcolsep}{1.9pt}
    \resizebox{\textwidth}{!}{
        \begin{tabular}{lccccccc}
\toprule
\textbf{Model} & \textbf{PIQA}$\uparrow$ & \textbf{ARC-e}$\uparrow$ & \textbf{ARC-c}$\uparrow$ & \textbf{HellaSwag}$\uparrow$ & \textbf{Winogrande}$\uparrow$ & \textbf{MMLU}$\uparrow$ & \textbf{Avg.}$\uparrow$ \\
    \midrule
    \textbf{T}: \qwenTeacher\citep{yu2025rlpr} & $78.7$ & $80.3$ & $\mathbf{63.7}$ & $\mathbf{80.2}$ & $\mathbf{76.4}$ & $\mathbf{75.3}$ & $\mathbf{75.8}$ \\
    \textbf{S}: \baselineQrwkvBase \citep{goldstein2025radlads} & $\mathbf{79.4}$ & $\mathbf{80.6}$ & $60.2$ & $77.9$ & $73.9$ & $65.4$ & $72.9$ \\

    \midrule
    \textbf{T}: \llamaTeacher \citep{grattafiori2024llama3} & $\mathbf{80.1}$ & $81.6$ & $\mathbf{57.8}$ & $81.9$ & $76.9$ & $65.9$ & $74.0$ \\
    \textbf{S}: \baselineLolcats \citep{zhang2025lolcats} & $79.5$ & $\mathbf{83.2}$ & $54.6$ & $75.3$ & $74.0$ & $52.9$ & $69.9$ \\
    \textbf{S}: \llamaStudent & $80.0$ & $82.9$ & $55.9$ & $\mathbf{82.3}$ & $\mathbf{79.0}$ & $\mathbf{66.1}$ & $\mathbf{74.4}$ \\

    \midrule
    \textbf{T}: \olmoTeacher \cite{olmo2025olmo3} & $\mathbf{78.1}$ & $80.7$ & $\mathbf{58.4}$ & $56.7$ & $\mathbf{73.4}$ & $\mathbf{66.4}$ & $\mathbf{69.0}$ \\
    \textbf{S}: \olmoStudent & $77.9$ & $\mathbf{81.1}$ & $57.8$ & $\mathbf{56.9}$ & $72.8$ & $65.7$ & $68.7$ \\

    \specialrule{1.5pt}{0.3ex}{0.3ex}
    \textbf{T}: \qwenTeacherIT \citep{yu2025rlpr} & $74.5$ & $68.6$ & $59.4$ & $74.5$ & $64.2$ & $\mathbf{74.5}$ & $69.3$ \\
    \textbf{S}: \baselineQrwkv \citep{goldstein2025radlads} & $76.9$ & $74.9$ & $\mathbf{62.0}$ & $\mathbf{78.1}$ & $69.9$ & $68.2$ & $\mathbf{71.7}$ \\
    \textbf{S:} \qwenStudentIT & $\mathbf{79.4}$ & $\mathbf{80.2}$ & $60.8$ & $60.0$ & $74.7$ & $73.7$ & $71.4$ \\
    \textbf{S:} \qwenStudentITGeneralist & $79.1$ & $73.7$ & $54.1$ & $58.9$ & $\mathbf{75.4}$ & $66.6$ & $68.7$ \\
    
    \midrule
    \textbf{T}: \llamaThreeTeacherIT \citep{grattafiori2024llama3} & $77.3$ & $\mathbf{76.9}$ & $\mathbf{66.2}$ & $\mathbf{72.4}$ & $\mathbf{73.9}$ & $\mathbf{68.6}$ & $\mathbf{72.6}$ \\
    \textbf{S}: \baselineMambaLlama \citep{wang2024mamba} & $\mathbf{82.8}$ & $74.8$ & $62.1$ & $64.1$ & $59.2$ & $56.8$ & $66.6$ \\
    
    \midrule
    \textbf{T}: \llamaTeacherIT \citep{grattafiori2024llama3} & $\mathbf{79.9}$ & $\mathbf{81.8}$ & $56.7$ & $\mathbf{59.3}$ & $\mathbf{78.0}$ & $68.9$ & $70.1$ \\
    \textbf{S}: \llamaStudentIT & $79.8$ & $81.2$ & $56.0$ & $58.7$ & $76.9$ & $68.0$ & $\mathbf{71.4}$ \\
    \textbf{S}: \llamaStudentITGeneralist & $78.8$ & $71.1$ & $\mathbf{58.4}$ & $57.9$ & $73.2$ & $\mathbf{72.8}$ & $67.9$ \\

\bottomrule
\end{tabular}

    }
    \label{tab:language-understanding}
\end{table}

\textbf{Language Generation \& Reasoning.}
Similarly, we report the raw performance scores across all language generation and reasoning tasks for all student and teacher models in Table~\ref{tab:language-generation}.
Again, we observe that our distilled students almost match their respective teacher performances, while alternative distillation recipes fall short.

\begin{table}
    \centering
    \caption{
        Raw scores for \textbf{downstream evaluations} on \textbf{language generation tasks} tasks.
        Our models perform comparably or slightly exceed their respective \llamaTeacher and \olmoTeacher teacher models, while significantly outperforming distilled models with alternative linearization recipes.
    }
    \label{tab:language-generation}
    \setlength{\tabcolsep}{1.9pt}
    \resizebox{0.7\textwidth}{!}{
        \begin{tabular}{lcccccccc}
\textbf{Model} &
\rotatebox{90}{\textbf{GPQA-D (0)}$\uparrow$} &
\rotatebox{90}{\textbf{GPQA (0)}$\uparrow$} &
\rotatebox{90}{\textbf{GSM8K (8)}$\uparrow$} &
\rotatebox{90}{\textbf{HumanEval (0)}$\uparrow$} &
\rotatebox{90}{\textbf{HumanEval+ (0)}$\uparrow$} &
\rotatebox{90}{\textbf{MBPP (3)}$\uparrow$} &
\rotatebox{90}{\textbf{MBPP+ (3)}$\uparrow$} &                   
\rotatebox{90}{\textbf{Avg.}$\uparrow$} \\
\specialrule{1.5pt}{0.3ex}{0.3ex}
\textbf{T}: \qwenTeacher \citep{yu2025rlpr} & $\mathbf{32.3}$ & $\mathbf{30.8}$ & $\mathbf{80.8}$ & $\mathbf{65.9}$ & $\mathbf{47.6}$ & $\mathbf{62.6}$ & $\mathbf{80.4}$ & $\mathbf{57.2}$ \\
\textbf{S}: \baselineQrwkvBase \citep{goldstein2025radlads} & $9.60$ & $13.2$ & $23.8$ & $20.7$ & $33.5$ & $0.0$ & $27.0$ & $14.4$ \\

\midrule
\textbf{T}: \llamaTeacher \citep{grattafiori2024llama3} & $10.6$ & $13.8$ & $48.4$ & $35.4$ & $\mathbf{29.9}$ & $\mathbf{48.4}$ & $\mathbf{61.9}$ & $35.5$ \\ 
\textbf{S}: \baselineLolcats \citep{zhang2025lolcats} & $1.77$ & $2.23$ & $3.87$ & $2.13$ & $3.35$ & $0.0$ & $0.0$ & $1.9$ \\ 
\textbf{S}: \llamaStudent & $\mathbf{17.7}$ & $\mathbf{15.8}$ & $\mathbf{57.8}$ & $\mathbf{39.0}$ & $23.8$ & $42.8$ & $56.9$ & $\mathbf{36.3}$ \\

\midrule
\textbf{T}: \olmoTeacher \cite{olmo2025olmo3} & $\mathbf{20.7}$ & $19.0$ & $\mathbf{67.5}$ & $\mathbf{32.9}$ & $\mathbf{28.7}$ & $\mathbf{50.6}$ & $\mathbf{71.7}$ & $\mathbf{41.6}$ \\
\textbf{S}: \olmoStudent & $17.7$ & $\mathbf{19.4}$ & $56.6$ & $32.3$ & $27.4$ & $36.8$ & $52.4$ & $34.7$ \\

\bottomrule

\end{tabular}

    }
\end{table}

\begin{table}
   \centering
   \caption{
       Raw scores for \textbf{downstream evaluations} on \textbf{language generation tasks} tasks.
       Our models perform comparably to their respective \llamaTeacherIT and \qwenTeacherIT~teacher models, while significantly outperforming distilled models with alternative linearization recipes.
   }
   \label{tab:language-generation-it}
   \setlength{\tabcolsep}{1.9pt}
   \resizebox{\textwidth}{!}{
       \begin{threeparttable}
\begin{tabular}{lccccccccccccccc}
\textbf{Model} &
\rotatebox{90}{\textbf{GSM8K (8)}$\uparrow$} &
\rotatebox{90}{\textbf{MATH500 (0)}$\uparrow$} &
\rotatebox{90}{\textbf{MATH (0)}$\uparrow$} &
\rotatebox{90}{\textbf{MATH Level 5 (0)}$\uparrow$} &
\rotatebox{90}{\textbf{HumanEval (0)}$\uparrow$} &
\rotatebox{90}{\textbf{HumanEval+ (0)}$\uparrow$} &
\rotatebox{90}{\textbf{MBPP (3)}$\uparrow$} &
\rotatebox{90}{\textbf{MBPP+ (3)}$\uparrow$} &  
\rotatebox{90}{\textbf{CruxEval-O (0)}$\uparrow$} &
\rotatebox{90}{\textbf{CruxEval-I (0)}$\uparrow$} &  
\rotatebox{90}{\textbf{GPQA-D (0)}$\uparrow$} &
\rotatebox{90}{\textbf{GPQA (0)}$\uparrow$} &     
\rotatebox{90}{\textbf{IfEval (0)}$\uparrow$} & 
\rotatebox{90}{\textbf{MT-Bench (0)}$\uparrow$} &     
\rotatebox{90}{\textbf{Avg.}$\uparrow$} \\
\specialrule{1.5pt}{0.3ex}{0.3ex}

\textbf{T}: \qwenTeacherIT \citep{yu2025rlpr} & $\mathbf{0.90}$ & $\mathbf{0.74}$ & $\mathbf{0.74}$ & $\mathbf{0.49}$ & $\mathbf{0.81}$ & $\mathbf{0.74}$ & $0.61$ & $\mathbf{0.79}$ & $0.41$ & $0.41$ & $\mathbf{0.34}$ & $\mathbf{0.36}$ & $\mathbf{0.74}$ & $5.20$ & $0.95$ \\
\textbf{S}: \baselineQrwkv \citep{goldstein2025radlads} & $0.53$ & $0.40$ & $0.39$ & $0.12$ & $0.62$ & $0.55$ & $0.49$ & $0.69$ & $0.18$ & $0.22$ & $0.22$ & $0.19$ & $0.59$ & $3.96$ & $0.65$ \\
\textbf{S:} \qwenStudentIT & $\mathbf{0.90}$ & $0.65$ & $0.66$ & $0.42$ & $0.78$ & $0.71$ & $\mathbf{0.63}$ & $\mathbf{0.79}$ & $\mathbf{0.42}$ & $\mathbf{0.42}$ & $0.26$ & $0.22$ & $0.67$ & $\mathbf{5.96}$ & $\mathbf{0.96}$ \\
\textbf{S:} \qwenStudentITGeneralist & $0.89$ & $0.63$ & $0.62$ & $0.36$ & $0.76$ & $0.68$ & $0.62$ & $0.78$ & $0.37$ & $0.36$ & $0.19$ & $0.14$ & $0.62$ & $5.55$ & $0.90$ \\
\midrule
\textbf{T}: \llamaTeacherIT \citep{grattafiori2024llama3} & $\mathbf{0.85}$ & $0.51$ & $0.50$ & $0.19$ & $\mathbf{0.63}$ & $\mathbf{0.57}$ & $\mathbf{0.59}$ & $0.69$ & $0.26$ & $\mathbf{0.28}$ & $\mathbf{0.30}$ & $\mathbf{0.32}$ & $\mathbf{0.78}$ & $5.08$ & $0.83$ \\
\textbf{S}: \llamaStudentIT & $0.83$ & $\mathbf{0.54}$ & $\mathbf{0.55}$ & $\mathbf{0.22}$ & $\mathbf{0.63}$ & $0.56$ & $0.54$ & $0.66$ & $\mathbf{0.34}$ & $0.27$ & $0.21$ & $0.20$ & $0.69$ & $\mathbf{6.05}$ & $\mathbf{0.88}$ \\
\textbf{S}: \llamaStudentITGeneralist & $0.82$ & $0.37$ & $0.37$ & $\mathbf{0.22}$ & $0.50$ & $0.47$ & $0.54$ & $\mathbf{0.71}$ & $0.29$ & $0.25$ & $0.16$ & $0.14$ & $0.49$ & $5.45$ & $0.77$ \\
\midrule
\textbf{T}: \llamaThreeTeacherIT \citep{grattafiori2024llama3} & $\mathbf{0.82}$ & $\mathbf{0.29}$ & $\mathbf{0.29}$ & $\mathbf{0.09}$ & $\mathbf{0.57}$ & $\mathbf{0.54}$ & $\mathbf{0.56}$ & $\mathbf{0.74}$ & $\mathbf{0.26}$ & $\mathbf{0.20}$ & $\mathbf{0.34}$ & $\mathbf{0.29}$ & $\mathbf{0.77}$ & $\mathbf{5.50}$ & $\mathbf{0.80}$ \\
\textbf{S}: \baselineMambaLlama \citep{wang2024mamba} & $0.68$ & $0.12$ & $0.12$ & $0.04$ & $0.38$ & $0.34$ & $0.33$ & $0.41$ & $0.09$ & $0.16$ & $0.25$ & $0.22$ & $0.52$ & $3.97$ & $0.54$ \\

\end{tabular}

\begin{tablenotes}
\footnotesize
\item[*]50\% attention layers
\end{tablenotes}
\end{threeparttable}

   }
\end{table}

\begin{figure*}
    \centering   
    \includegraphics[width=0.49\linewidth]{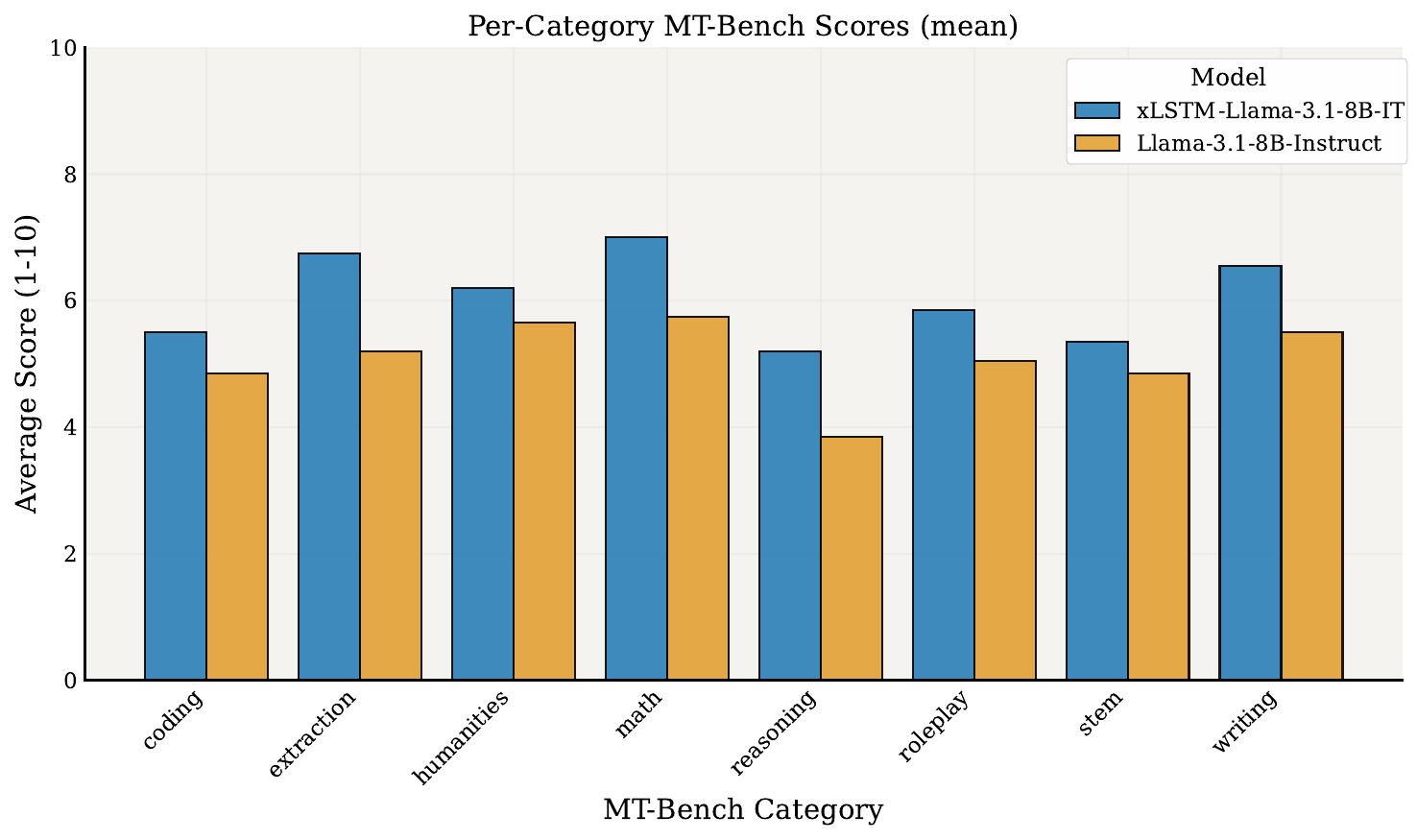}
    \includegraphics[width=0.49\linewidth]{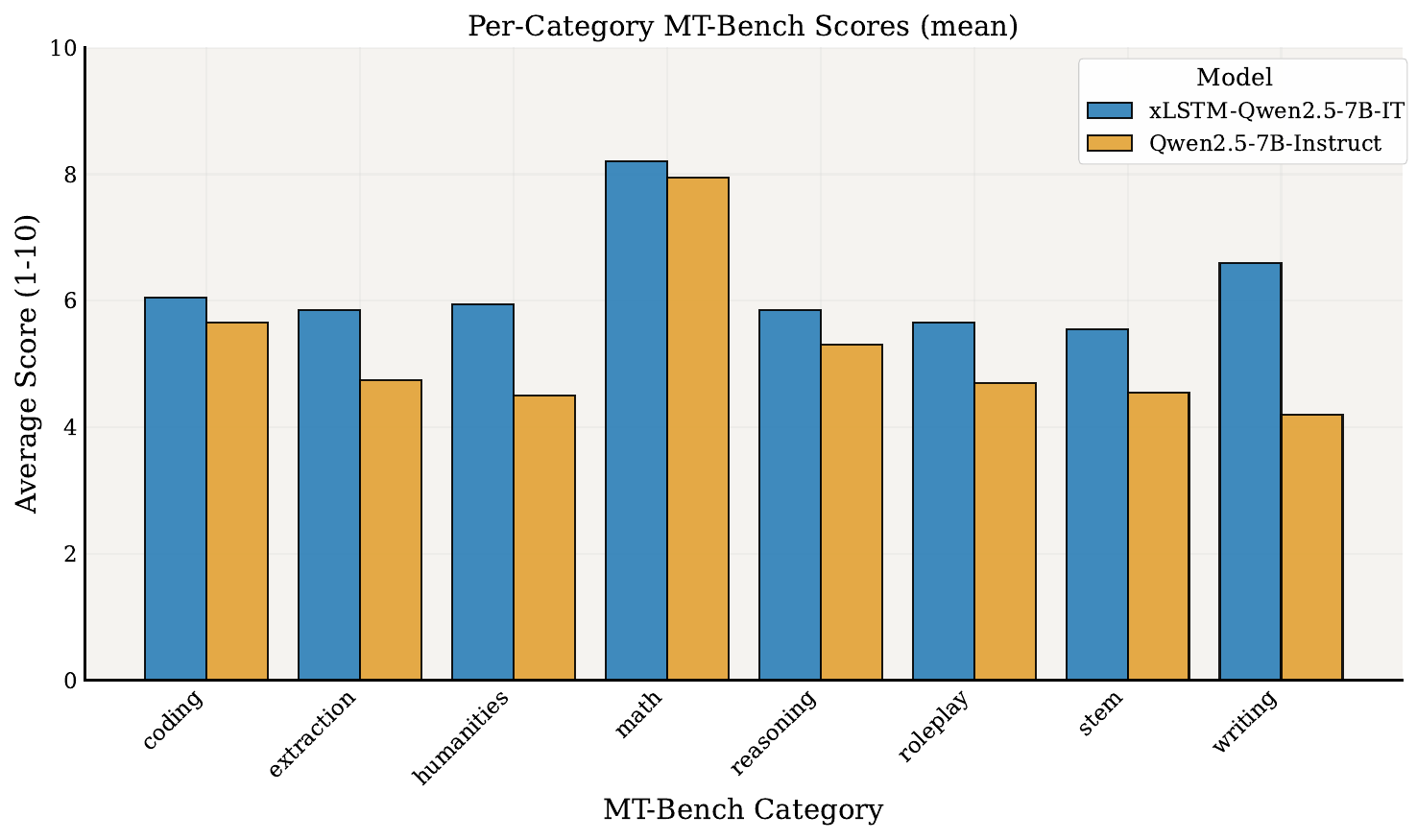}
    \caption{MT-Bench performance per category as judged by GPT-5.1.}
    \label{fig:mtbench}
\end{figure*}

\textbf{MT-bench.}
Figure \ref{fig:mtbench} shows MT-Bench \citep{zheng2023judging} performance as evaluated by GPT5.1 broken down by category.
Both \qwenStudentIT and \llamaStudentIT outperform their teacher models on all 7 MT-bench categories.

\textbf{Needle in a Haystack.}
In Appendix Table~\ref{tab:niah}, we report Needle-in-a-Haystack (NIAH; \cite{hsieh2024ruler}) accuracy for both the single-needle and multi-needle variants across four context lengths (1k, 4k, 8k, and 16k tokens). Overall, the instruction-tuned baselines (\llamaTeacherIT~and \qwenTeacherIT) maintain near-perfect recall across all lengths, whereas our distilled Students degrade with increasing context, with the largest drop occurring at 4k and compounding further at 8k and 16k. We note that we also observe substantial long-context degradation when continually fine-tuning the Transformer baseline without linearization, and because our Student was not exposed to long-context distillation data during adaptation, it remains unclear whether the sharp decline reflects an inherent memory limitation of the fixed-size state or could be partially mitigated with improved long-context instruction tuning; we leave this question to future work.

\begin{table}
    \centering
    \caption{Comparison of \llamaTeacherIT~and \qwenTeacherIT~against our Students on Needle-in-a-Haystack tasks.}
    \small
\setlength{\tabcolsep}{4pt}
\begin{tabular}{lcccc|cccc}
\toprule
Model & \multicolumn{4}{c|}{NIAH Single} & \multicolumn{4}{c}{NIAH Multi} \\
\cmidrule(lr){2-5}\cmidrule(lr){6-9}
& $\mathbf{1024}$ & $\mathbf{4096}$ & $\mathbf{8192}$ & $\mathbf{16384}$ & $\mathbf{1024}$ & $\mathbf{4096}$ & $\mathbf{8192}$ & $\mathbf{16384}$ \\
\midrule
\llamaTeacherIT     & $1.000$ & $1.000$ & $1.000$ & $1.000$ & $1.000$ & $1.000$ & $1.000$ & $1.000$ \\
\llamaStudentIT        & $0.686$ & $0.148$ & $0.078$ & $0.024$ & $0.664$ & $0.130$ & $0.088$ & $0.034$ \\
\midrule
\qwenTeacherIT       & $0.998$ & $0.998$ & $1.000$ & $1.000$ & $1.000$ & $1.000$ & $1.000$ & $0.998$ \\
\qwenStudentIT         & $0.656$ & $0.140$ & $0.078$ & $0.024$ & $0.640$ & $0.116$ & $0.088$ & $0.034$ \\
\bottomrule
\end{tabular}

    \label{tab:niah}
\end{table}

\subsection{Inference Time Analysis}

In this section, we provide additional details on our inference time tests. 

\textbf{Setup.} 
We run all our inference time tests on a single H100 GPU with 80GB~of memory. 
Our implementations for both student and teacher are based on the \texttt{transformers} library \citet{wolf2020transformers} and their respective classes for Llama~3 \citep{touvron2023llama}.
For our hybrid student model, we replace the self-attention mechanism of the teacher model with our hybrid mechanism of mLSTM and \ac{swa}.
To accelerate runtimes, we leverage \texttt{torch.compile} using a static KV-cache. 
For the Transformer-based teacher, we leverage FlashAttention-2 \citep{dao2024flashattention2}. 
Similarly, for our hybrid student, we make use of the Triton kernels released by \citet{beck2025tiled} for the mLSTM part and FlashAttention for the sliding window part.
To enable compilation of our hybrid student via \texttt{torch.compile}, we utilize a custom static cache implementation that retains both the mLSTM states and the relevant keys/values of sink tokens and the sliding window.

\begin{figure*}
    \centering
    \begin{subfigure}{.41\linewidth}
        \includegraphics[width=\linewidth]{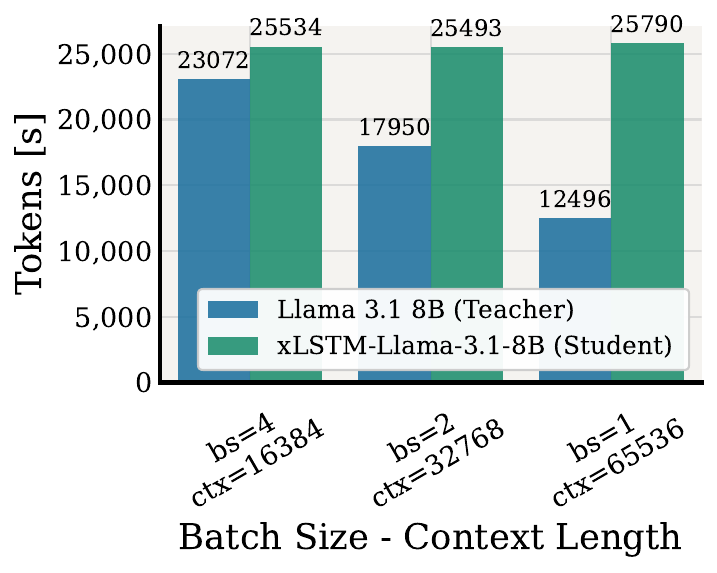}
        \caption{Prefill Throughput}
    \end{subfigure}%
    \begin{subfigure}{.45\linewidth}
        \includegraphics[width=\linewidth]{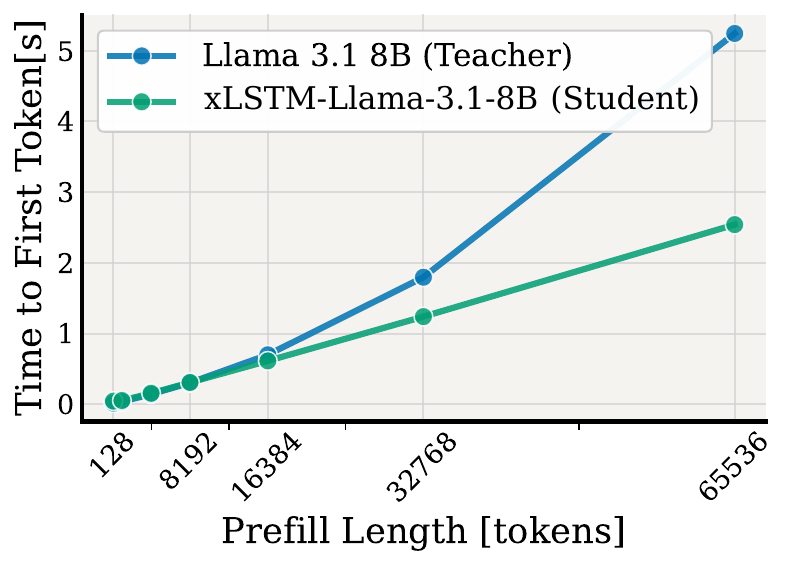}
        \caption{Prefill Latency, $B=1$}
    \end{subfigure}
    \caption{
        Inference comparison for the \textbf{prefilling stage} between the Transformer-based teacher and our mLSTM-based student.
        In \textbf{(a)}, we report prefill throughput for varying context lengths and batch sizes.
        In \textbf{(b)}, we show the prefilling latency for varying prefill lengths and $B=1$.
    }
    \label{fig:inf-comparison-prefill}
\end{figure*}

\begin{figure*}[!t]
    \centering
    \includegraphics[width=0.50\linewidth]{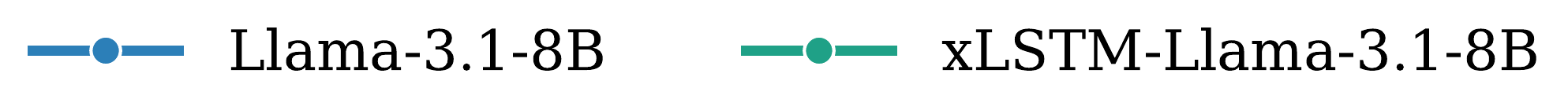}
    \begin{subfigure}{.3\linewidth}
        \includegraphics[width=\linewidth]{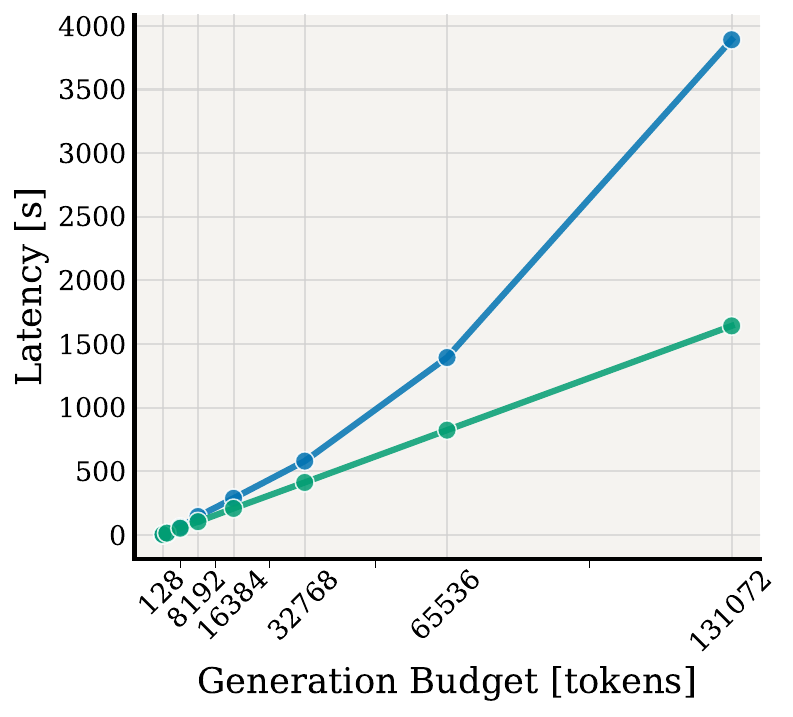}
        \caption{Latency, $B=1$}
    \end{subfigure}%
    \begin{subfigure}{.3\linewidth}
        \includegraphics[width=\linewidth]{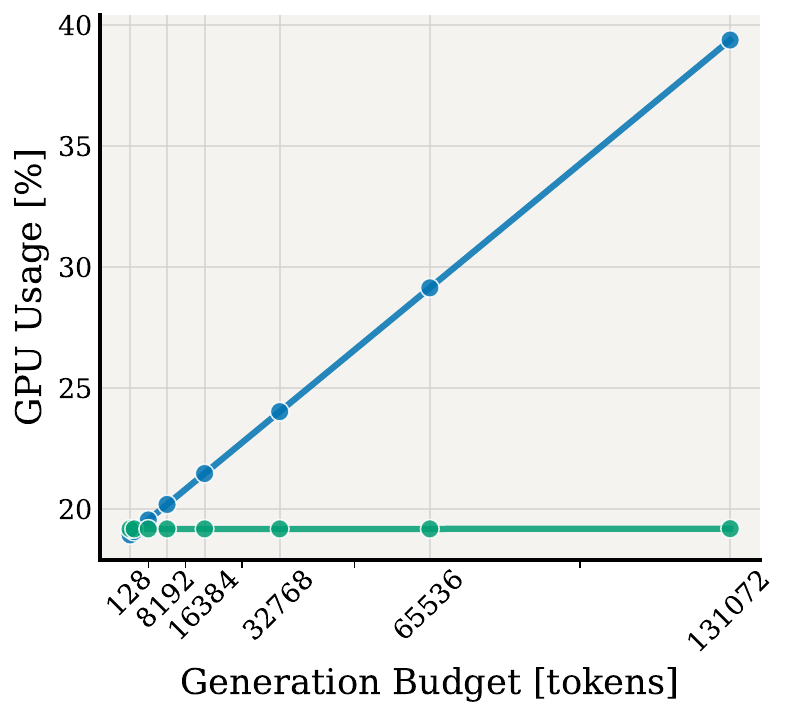}
        \caption{GPU RAM \%, $B=1$}
    \end{subfigure}%
    \begin{subfigure}{.3\linewidth}
        \includegraphics[width=\linewidth]{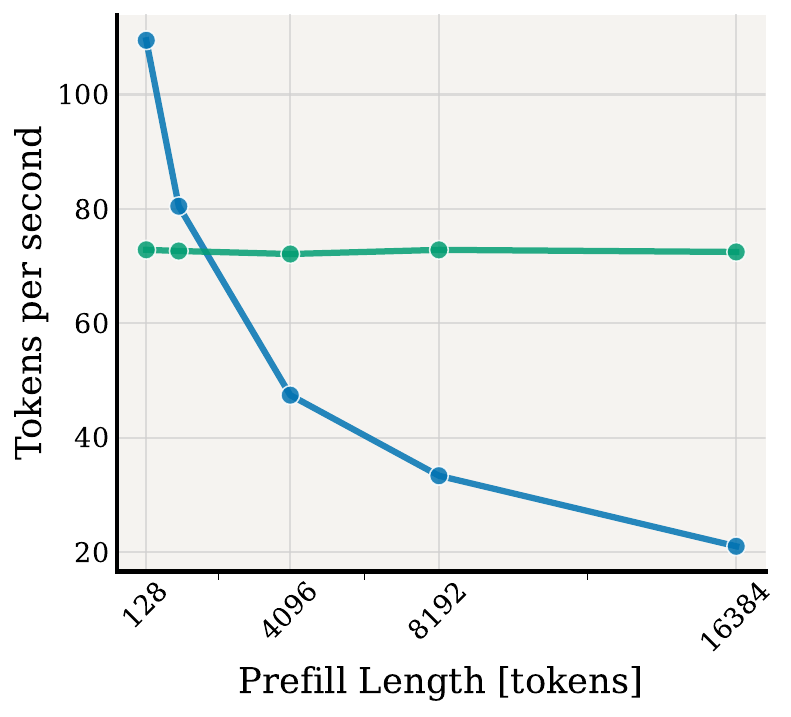}
        \caption{Throughput, $B=8$}
    \end{subfigure}
    \caption{
        Inference comparison for the \textbf{generation stage} between the Transformer-based teacher and our xLSTM-based student.
        In \textbf{(a)}, we show generation latency at different generation budgets ($B=1$).
        In \textbf{(b)}, we report the memory consumption in \% of GPU memory during the generation ($B=1)$.
        In \textbf{(c)}, we show the generation throughput when generating 100 tokens with varying prefill lengths and $B=8$.
    }
    \label{fig:inf-comparison-generation}
\end{figure*}

\textbf{Prefill vs. generation.}
We separate our inference time tests into two stages: prefilling and generation. 
While the prefilling stage encodes the input prompt by the user and populates the KV cache, the generation or decoding stage autoregressively samples tokens until sequence termination, starting from the pre-filled KV cache \citep{pope2023efficiently}.
Our inference tests are characterized by three core hyperparameters, the batch size~$B$, the context length~$C$, and the generation budget~$G$. 
Consequently, if one would only perform prefilling without generating any tokens, then $G=0$. 
Similarly, if we only perform generation without any prefill sequence, then $C=0$
These two scenarios are reflected in Figure~\ref{fig:inf-comparison-prefill}a and Figure~\ref{fig:inf-comparison-generation}a, respectively.
For every combination of $B$, $C$, and $G$, we always first conduct three warmup runs, which include the compilation of our model. 
Afterwards, we record runtimes, memory consumption, and throughput across five runs and average metrics across them.

\label{appendix:inference}
\begin{figure*}
    \centering
    \includegraphics[width=0.50\linewidth]{figures/inference/legend_twocol_corrGK.pdf}
    
    \includegraphics[width=1\linewidth]{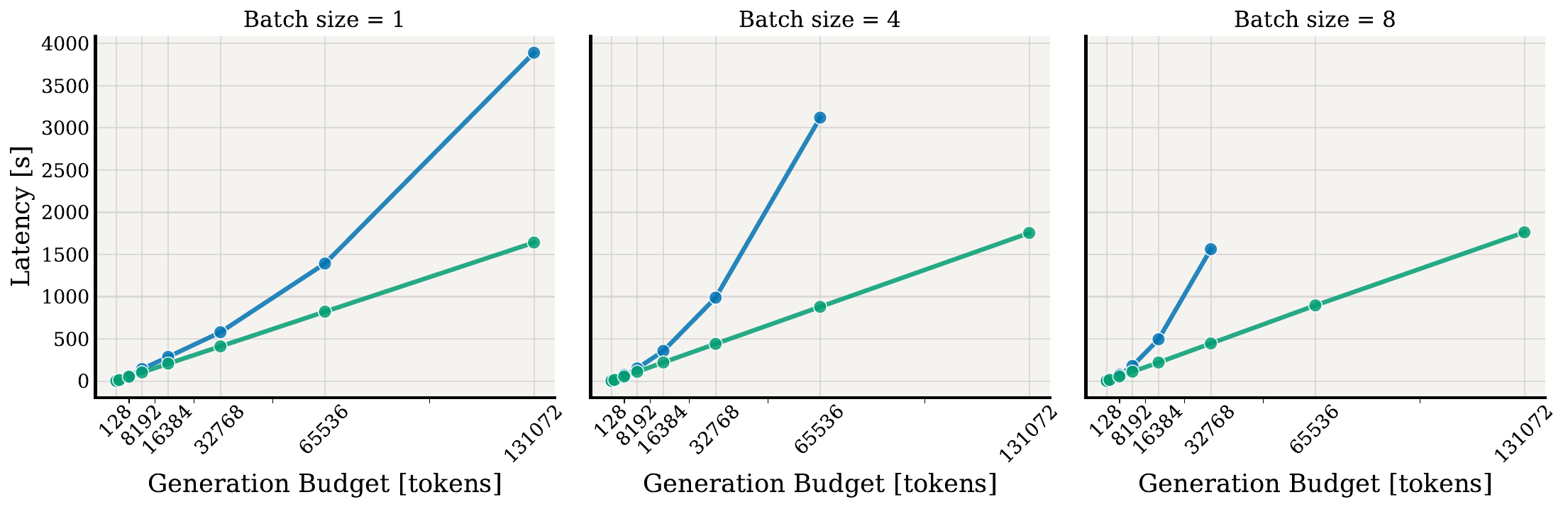}
    \caption{
        \textbf{Latency}.
        We report the latency for generation with varying token generation budgets and batch sizes.
        Our mLSTM-based student exhibits lower generation latency than the Transformer-based teacher.
        This advantage grows with larger generation budgets and batch sizes.
        Missing dots for the teacher indicate \acs{oom}.
    }
    \label{fig:inference-genlatency-bs}
\end{figure*}

\begin{figure*}
    \centering
    \includegraphics[width=0.50\linewidth]{figures/inference/legend_twocol_corrGK.pdf}
    
    \includegraphics[width=1\linewidth]{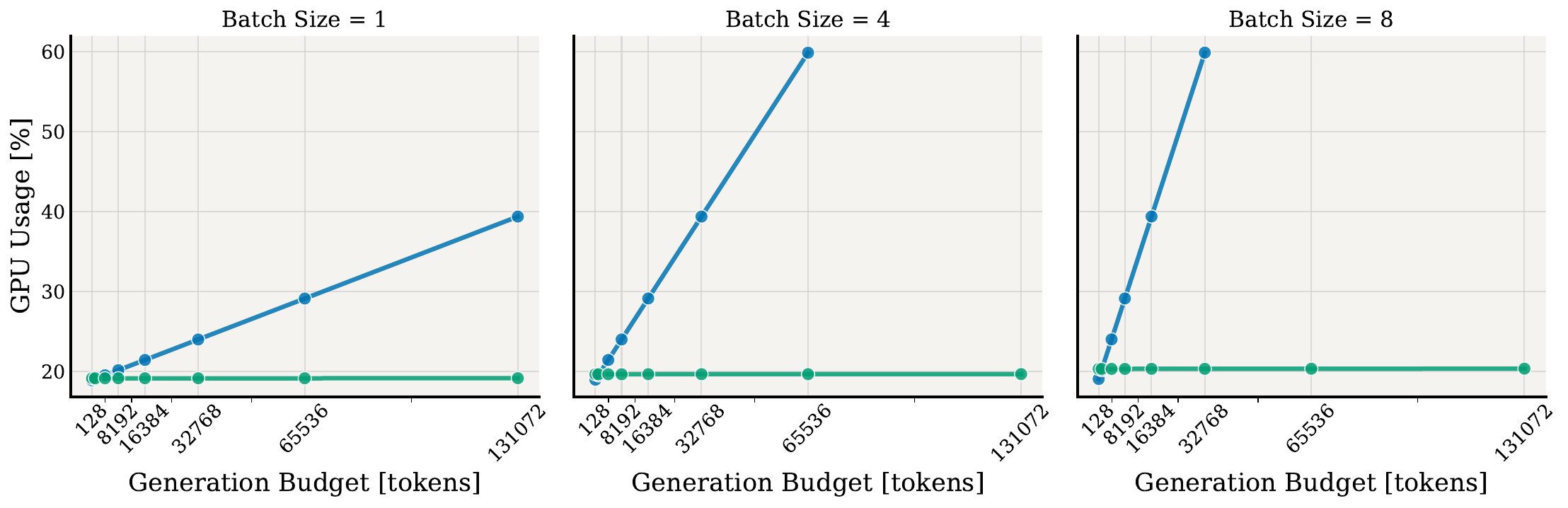}
    \caption{
        \textbf{GPU RAM}.
        We report the memory consumption in \% of GPU memory during the generation for varying batch sizes.
        Our mLSTM-based student requires significantly less memory compared to the Transformer-based teacher.
        This advantage grows with larger generation budgets and batch sizes.
        Missing dots for the teacher indicate \acs{oom}.
    }
    \label{fig:inference-genmem-bs}
\end{figure*}

\textbf{Generation latency \& memory consumption.}
To complement the results that we presented in Section \ref{sec:inference}, we report additional metrics for generation latency and memory consumption across varying batch sizes $B\in[1,4,8]$ and sequence lengths $C\in[128,1024,4096,16384,32768,65536,131072]$ in Figures~\ref{fig:inference-genlatency-bs} and~\ref{fig:inference-genmem-bs}, respectively. 
The purpose of this experiment is to better understand how the inference time advantages behave with increasing batch sizes and sequence lengths.

In Figure~\ref{fig:inference-genlatency-bs}, we make two important observations. 
First, our hybrid student only exhibits a slight increase in total inference latency when increasing the batch size from~1 to~8.
This is because its computational complexity does not grow with the sequence length, due to the recurrent inference mode of mLSTM and the fixed sliding window of 256~tokens.
Consequently, for the largest batch size we compare ($B=8$), the computation remains memory-bound.
Second, we observe that the computational demand of the Transformer-based teacher grows faster with increasing sequence length, due to the quadratic complexity of self-attention. 
For example, when increasing the batch size from~1 to~4, the runtime at $C=65K$ of our teacher model increases two-fold.  
Similarly, the required memory grows quickly as the batch size increases and the KV cache gets larger, causing the model to \ac{oom}, as indicated by missing dots for the teacher.
The differences in RAM consumption are further highlighted in Figure~\ref{fig:inference-genmem-bs}. 
The RAM consumption of the Transformer-based teacher grows quickly with increasing sequence length and batch size. 
In contrast, our hybrid student remains constant along the sequence length and only exhibits a slight increase with larger batch sizes due to the constant memory complexity.   

\begin{figure*}
    \centering
    \includegraphics[width=0.50\linewidth]{figures/inference/legend_twocol_corrGK.pdf}
    
    \includegraphics[width=1\linewidth]{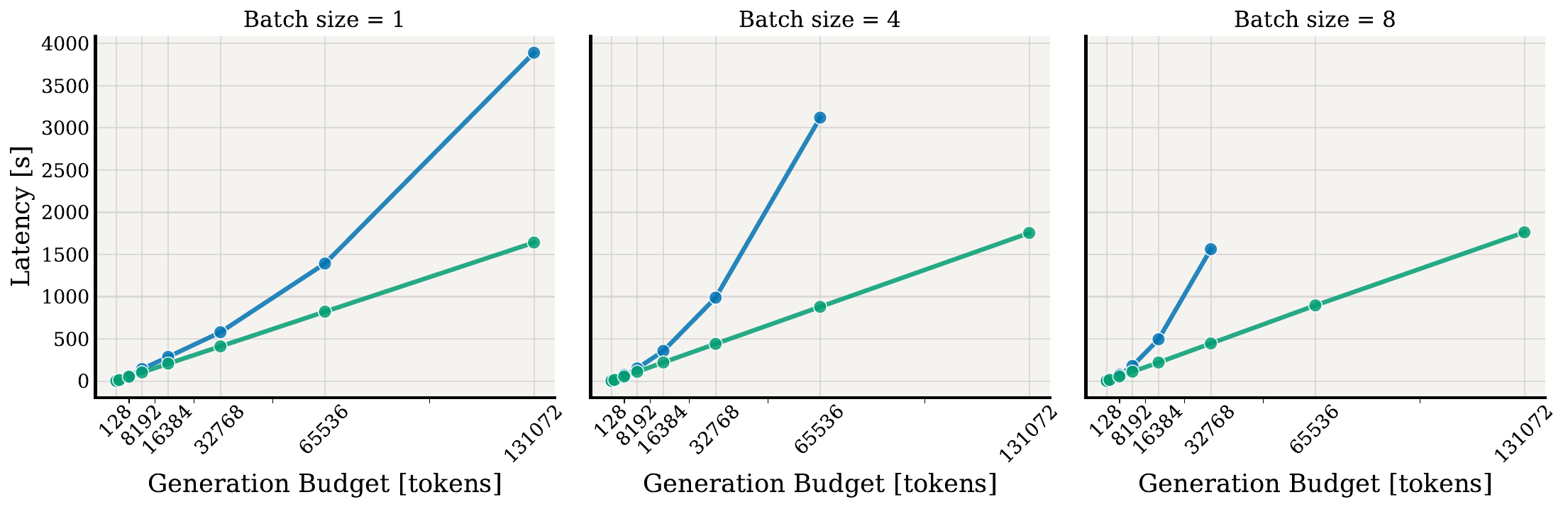}
    \caption{
        \textbf{Throughput}.
        We report the average throughput for generating 100~tokens with varying prefill lengths and batch sizes.
        Missing dots for the teacher indicate \acs{oom}.
    }
    \label{fig:inference-genthroughput-bs}
\end{figure*}

\textbf{Generation throughput}.
Finally, we report the average generation throughputs for a fixed generation budget of $G=100$ tokens with varying prefill lengths and $B \in [1,4,8]$ in Figure~\ref{fig:inference-genthroughput-bs}.
Again, we observe significantly higher throughputs for our hybrid student of up to almost~$4\times$ that of the Transformer-based teacher as the sequence length increases.
Note that due to OOMs, we only show metrics for $B=4$ and $B=8$ up to $C=32K$ and $C=16K$, respectively.

\section{Ablations}

In this section, we empirically analyse a variety of important components of our distillation recipe.

\subsection{Effects of SWA, Attention Sinks \& Gating}
\label{appendix-sec:architecture-ablation}

\begin{figure*}
    \centering   
    \includegraphics[width=0.83\linewidth]{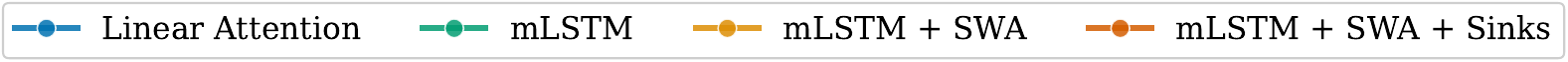}
    \begin{subfigure}{.45\linewidth}
        \includegraphics[width=\linewidth]{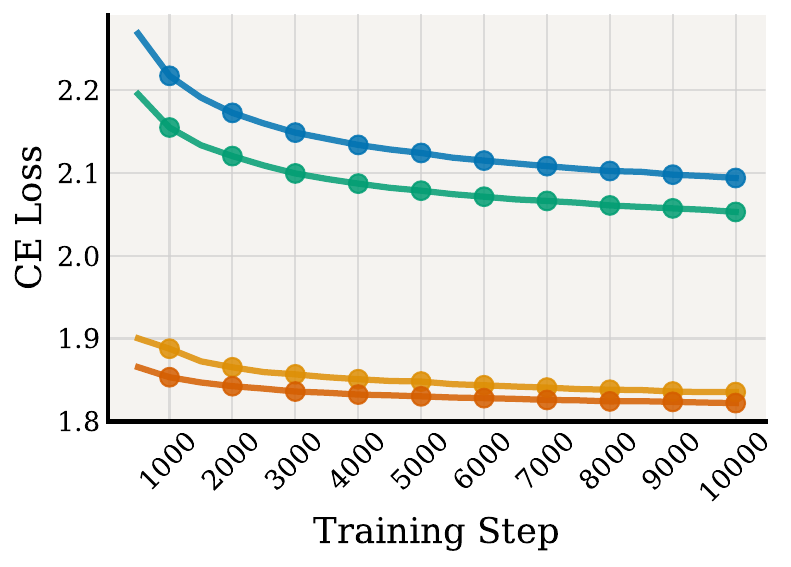}
    \end{subfigure}%
    \begin{subfigure}{.45\linewidth}
        \includegraphics[width=\linewidth]{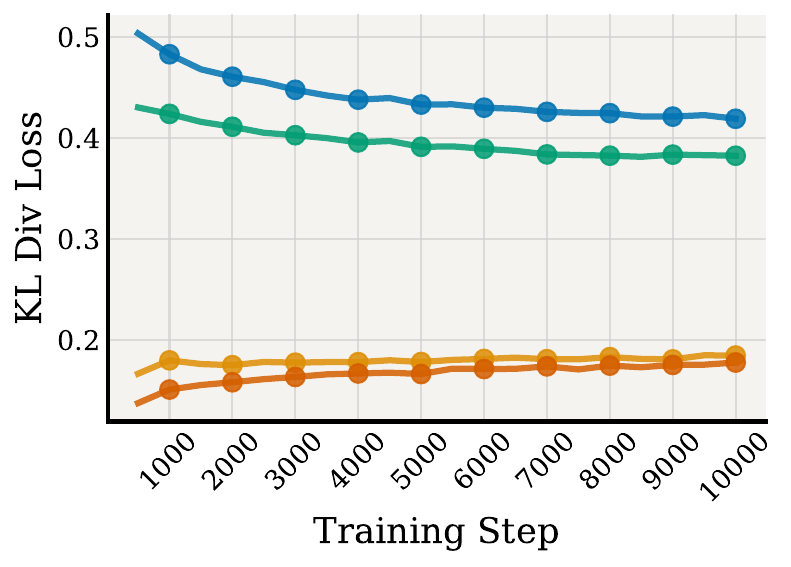}
    \end{subfigure}
    \caption{
        Ablation on the \textbf{effects of mLSTM, \ac{swa} \& sinks} on our hybrid student.
        We track the cross-entropy loss (left) and \ac{kl} loss (right) throughout stage~II and across individual components.
        All components contribute considerably to the final performance
    }
    \label{fig:component_ablations}
\end{figure*}

First, we ablate components of our hybrid attention operator under a fixed linearization recipe.
We compare four variants: (i) pure linear attention, (ii) mLSTM, (iii) mLSTM + \ac{swa}, and (iv) mLSTM + \ac{swa} + sink tokens.
For \ac{swa}, we use a window size $W=512$ and designate the first four tokens of each sequence as attention sinks.
We train with a \ac{ce}/\ac{kl} objective weighted by $\gamma=0.9$ and $\beta=0.1$.
Evaluation loss curves in Figure~\ref{fig:component_ablations} show that replacing linear attention with gated mLSTM reduces validation \ac{ce} throughout training, indicating that head-wise gates increase expressivity and better match the teacher.
Adding the \ac{swa} branch yields a further, uniform \ac{ce} reduction, consistent with exact short-range recall.
Introducing a small prefix of sink tokens provides an additional gain.
Overall, the full hybrid (mLSTM + \ac{swa} + sinks) converges faster and achieves the lowest \ac{ce}, i.e., the tightest student–teacher alignment.

\subsection{Effect of Distillation Objective}
\label{appendix-sec:distill-objective}

\begin{figure*}
    \centering   
    \includegraphics[width=0.75\linewidth]{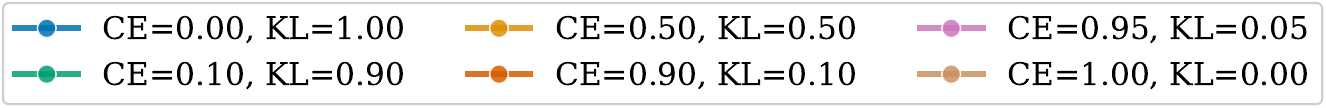}

    \begin{subfigure}{.45\linewidth}
        \includegraphics[width=\linewidth]{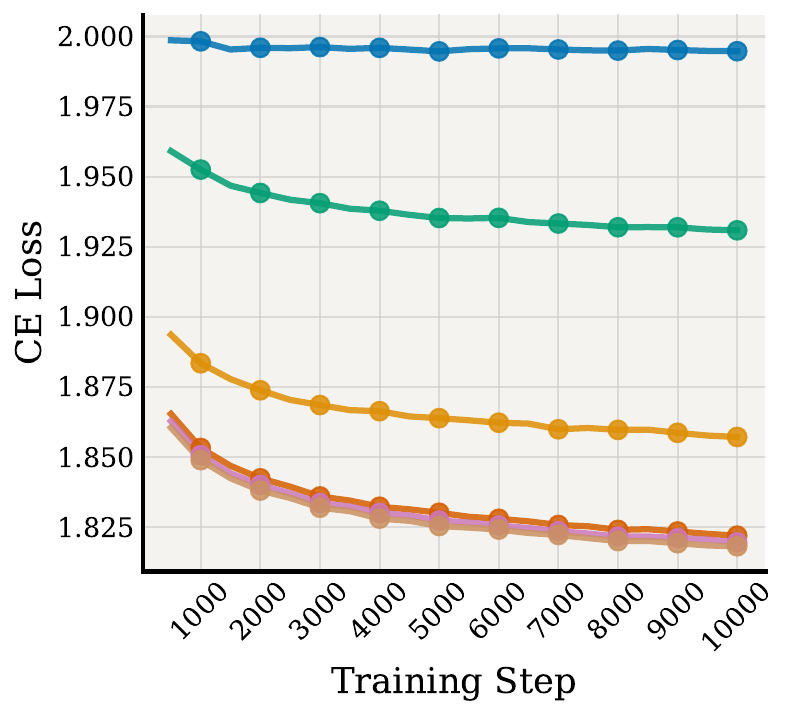}
    \end{subfigure}%
    \begin{subfigure}{.45\linewidth}
        \includegraphics[width=\linewidth]{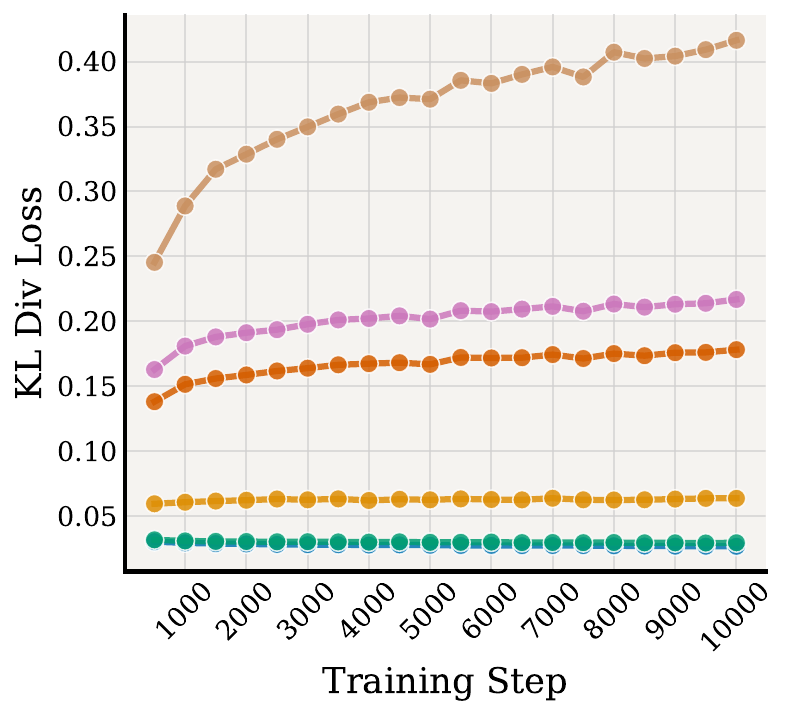}
    \end{subfigure}
    \caption{
        Ablation on the \textbf{effects of different loss weightings} on our hybrid student.
        We track the \ac{ce} (left) and \ac{kl} loss (right) throughout stage~II.
        We find that weighting \ac{ce} loss with~0.9 and \ac{kl} loss with~0.1 provides a good tradeoff between performance and teacher alignment.
    }
    \label{fig:loss_ablation}
\end{figure*}

To find the best tradeoff between \ac{ce} loss and the \ac{kl} between the student and teacher models, we sweep over \ac{ce} and \ac{kl} loss weights $\gamma$ and $\beta$ (see Eq.~\ref{eq:distill}).
Evaluation losses of the different fine-tuning configurations are shown in Figure~\ref{fig:loss_ablation}.
As the \ac{kl} $\beta$ grows, validation \ac{ce} rises and the student under-adapts.
As $\beta \to 0$, \ac{ce} is lowest but the student drifts, with \ac{kl} diverging.
Prior work observes that large post-finetuning \ac{kl} correlates with forgetting of capabilities \citep{shenfeld2025rl}.
Based on this, we adopt \ac{ce} $\gamma$=0.9 and \ac{kl} $\beta=0.1$.
This setting achieves \ac{ce} essentially matching the $\gamma=1.0, \beta=0$ configuration while keeping \ac{kl} dramatically smaller, providing substantial freedom to adapt to the new attention operators without sacrificing teacher alignment.
Notably, even a small \ac{kl} term materially improves alignment.
Therefore, we use \ac{ce} $\gamma=0.9$ and \ac{kl} $\beta=0.1$ as our base setting.

\subsection{PEFT vs. FFT}
\label{appendix-sec:peft-vs-fft}

\begin{figure*}
    \centering   
    \includegraphics[width=0.45\linewidth]{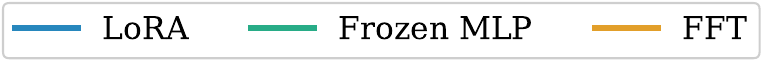}
    \begin{subfigure}{.45\linewidth}
        \includegraphics[width=\linewidth]{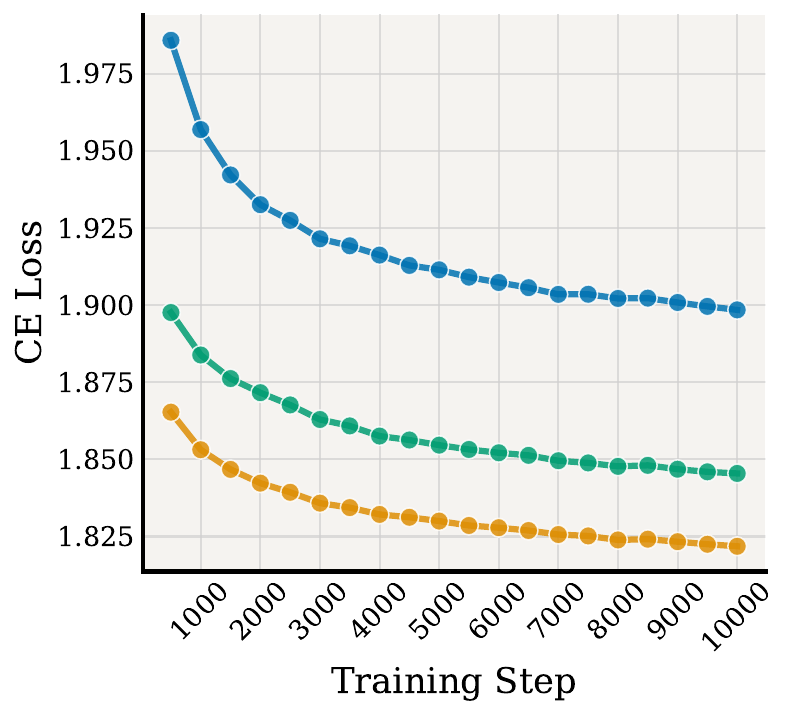}
    \end{subfigure}%
    \begin{subfigure}{.45\linewidth}
        \includegraphics[width=\linewidth]{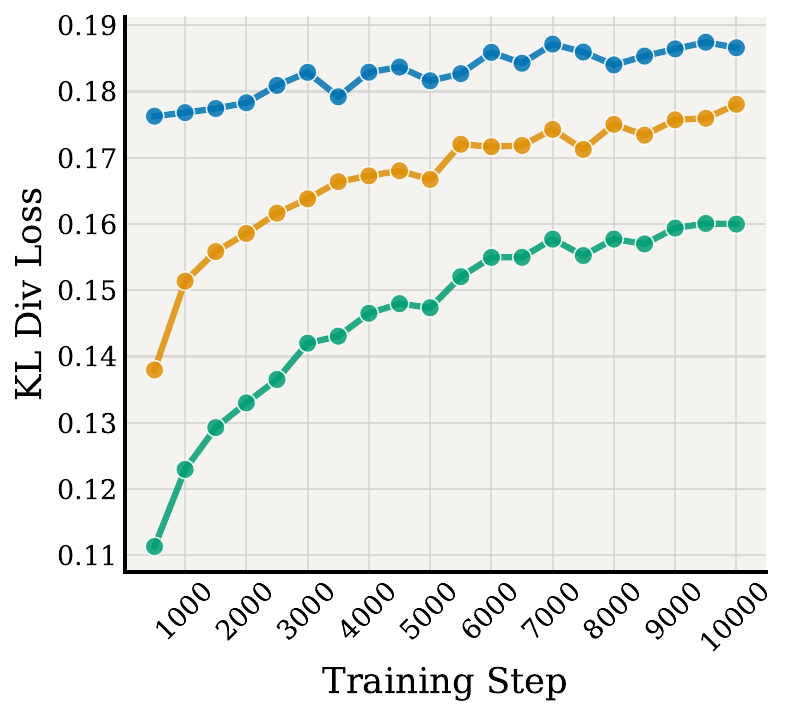}
    \end{subfigure}
    \caption{
        Ablation on the \textbf{effects of different training strategies} on our hybrid student.
        We track the \ac{ce} loss (left) and \ac{kl} loss (right) throughout stage~II.
        This analysis reveals that full finetuning achieves a comparable \ac{kl} while yielding superior downstream performance.
    }
    \label{fig:recipe_ablation}
\end{figure*}

Prior linearization recipes use low-rank adaptation (LoRA, \citealp{hu2022lora}) to recover performance lost during conversion \citep{zhang2025lolcats,nguyen2025lizard}.
While \ac{lora} is attractive for its cost and scalability, it is unclear whether its capacity suffices to close the student-teacher gap.
We therefore ablate three strategies spanning the efficiency–expressivity trade-off: (i) \ac{lora} with high ranks ($r=256$), (ii) updating only the sequence-mixer parameters while freezing all \ac{mlp} blocks and embeddings, and (iii) \ac{fft}.
We follow the baseline linearization setup in Section~\ref{sec:method} and train with \ac{ce} weight $\gamma = 0.9$ and \ac{kl} weight $\beta = 0.1$.
Figure~\ref{fig:recipe_ablation} reports validation \ac{ce} and \ac{kl}.
As expected, \ac{ce} decreases as more parameters are unfrozen: \ac{fft} achieves the lowest \ac{ce}, followed by mixer-only, then \ac{lora}.
Surprisingly, this additional flexibility does not increase deviation from the teacher.
A small \ac{kl} penalty is sufficient to keep all three methods comparably close in \ac{kl}.
Consequently, we adopt \ac{fft}, which offers the greatest capacity to adapt to the new attention operators while remaining close to the teacher model.

\subsection{Effect of Phases~I\&II}
\label{appendix-sec:phase12}

\noindent
\begin{minipage}[t]{0.54\linewidth}
\vspace{0pt}
We find that layer-wise hidden-state alignment (Phase~I) is necessary but not sufficient to recover most of the teacher's performance within a limited training budget.
This finding aligns with previous work by \citep{zhang2025lolcats}.
Consequently, a two-phase approach, hidden-state alignment followed by full finetuning (Phase~II), consistently outperforms standard finetuning given the same budget.
While Phase~I is crucial for aligning the student's intermediate hidden representations with the teacher's, the performance gains from this stage alone quickly plateau after a limited number of steps (Figure~\ref{fig:phase12_ablation}).
\end{minipage}\hfill
\begin{minipage}[t]{0.42\linewidth}
\vspace{0pt}
    \centering
    \refstepcounter{figure}
    \label{fig:phase12_ablation}
    \includegraphics[width=\linewidth]{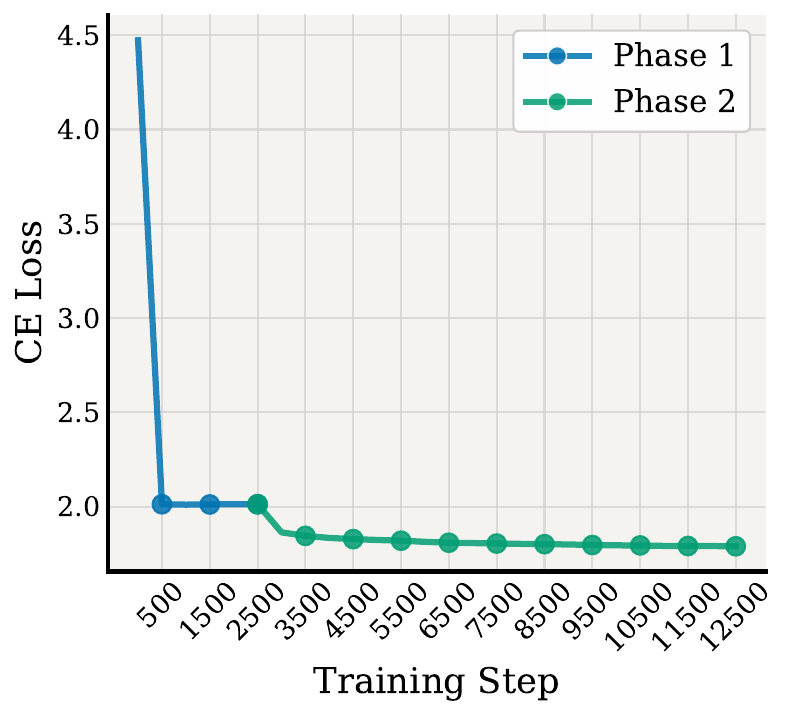}

    \smallskip
    \small\textbf{Figure~\thefigure.} Phase~I\&II ablation.
\end{minipage}

\end{document}